\documentclass[authoryear,preprint,12pt]{elsarticle}

\let\today\relax
\makeatletter
\def\ps@pprintTitle{%
    \let\@oddhead\@empty
    \let\@evenhead\@empty
    \def\@oddfoot{\footnotesize\itshape
         {Submitted Preprint} \hfill\today}%
    \let\@evenfoot\@oddfoot
    }
\makeatother

\usepackage{xcolor}
\usepackage{color,soul}

\usepackage{amssymb}
\usepackage{subcaption}

\usepackage{url}

\usepackage{booktabs} 
\usepackage{multirow}

\journal{Expert Systems with Applications}

\begin{document}

\begin{frontmatter}


\author[1]{Rrubaa Panchendrarajan\corref{cor1}}
\ead{r.panchendrarajan@qmul.ac.uk}

\author[1]{Arkaitz Zubiaga}
\ead{a.zubiaga@qmul.ac.uk}
\ead[url]{www.zubiaga.org}

\cortext[cor1]{Corresponding author}

\title{Synergizing Machine Learning \& Symbolic Methods: A Survey on Hybrid Approaches to Natural Language Processing}

\affiliation[1]{organization={School of Electronic Engineering and Computer Science, Queen Mary University of London},
             addressline={327 Mile End Rd, Bethnal Green},
             city={London},
             postcode={E1 4NS},
             country={United Kingdom}}

\begin{abstract}
The advancement of machine learning and symbolic approaches have underscored their strengths and weaknesses in Natural Language Processing (NLP). While machine learning approaches are powerful in identifying patterns in data, they often fall short in learning commonsense and the factual knowledge required for the NLP tasks. Meanwhile, the symbolic methods excel in representing knowledge-rich data. However, they struggle to adapt dynamic data and generalize the knowledge. Bridging these two paradigms through hybrid approaches enables the alleviation of weaknesses in both while preserving their strengths. Recent studies extol the virtues of this union, showcasing promising results in a wide range of NLP tasks. In this paper, we present an overview of hybrid approaches used for NLP. Specifically, we delve into the state-of-the-art hybrid approaches used for a broad spectrum of NLP tasks requiring natural language understanding, generation, and reasoning. Furthermore, we discuss the existing resources available for hybrid approaches for NLP along with the challenges and future directions, offering a roadmap for future research avenues. 
\end{abstract}




\end{frontmatter}


\section{Introduction}
The field of machine learning has witnessed remarkable progress over the past few decades achieving human-level performance in various Natural Language Processing (NLP) tasks. Large language models, in particular, have garnered global attention in the last couple of years, captivating not only the NLP community but also integrating into the daily routines of numerous professionals \citep{min2023recent}. Meanwhile, symbolic methods have seen significant advancements in representing human cognitive capabilities and knowledge-rich data enabling it to be more interpretable and comprehensible \citep{dale2000symbolic}. However, these two main pillars of computer science research possess their own set of advantages and disadvantages. 

Machine learning approaches are stronger in learning patterns and relationships in data via optimization strategies. However, they often fall short in capturing and interpreting the factual knowledge required for most downstream NLP tasks. Instead, they attempt to mimic the facts and knowledge present in the training data \citep{pan2023unifying}. This hinders both traditional machine learning approaches and deep learning methods from achieving high performance in knowledge-intensive tasks. Further, recent studies \citep{petroni-etal-2019-language} have demonstrated that even robust large language models trained using vast amounts of training data and parameters suffer from hallucinations generating non-factual responses \citep{zhang2023siren}. This raises significant concerns regarding the trustworthiness of such expensive large language models. Similarly, symbolic methods are stronger in resembling human cognitive abilities by explicitly capturing the knowledge required for a task. On the other hand, symbolic methods have shown poor learning skills compared to machine learning approaches, especially in handling dynamic data and generalizing the knowledge beyond training data \citep{pan2023unifying}.    

Recent focus has been turned into bridging the gap between machine learning approaches and symbolic methods to overcome the limitations of both when applied independently while retaining their strengths \citep{zhu2023knowledge}. This hybrid approach enables the statistical methods to utilize knowledge-enriched input data to improve the inference with external knowledge and to produce interpretable results. At the same time, symbolic methods are empowered with statistical learning to incorporate semantic knowledge and generate generalized knowledge representations. In particular, the combination of deep learning and symbolic methods has paved the way for a new research era called Neuro-symbolic methods \citep{hamilton2022neuro} aiming to develop trustworthy and interpretable NLP solutions. Furthermore, recent studies \citep{sarker2021neuro} have shown, that neuro-symbolic solutions gain benefits in four key research aspects including interpretability, generalization to handle both small training data and out of distributions, and error recovery due to aggregated advantages of both deep learning and symbolic methods. This has drawn the attention of the artificial intelligence research community to develop effective hybrid solutions to solve various real-world problems. 

The field of NLP has also begun to embrace hybrid techniques to develop effective real-world solutions. Especially, the existence of a larger number of textual knowledge bases has aided the rapid adoption of hybrid techniques over the past few years. Further, the representation of natural language is inherently a symbolic representation, hence the emergence of this new field, \textit{Hybrid NLP}, is an unsurprising development within the field of NLP. Inspired by this rapidly progressing research paradigm, this survey presents an overview of the literature on hybrid approaches that combine machine learning techniques and symbolic methods for NLP. Specifically, we address the following research questions in this survey.
\begin{itemize}
    \item What are the hybrid techniques used in different NLP tasks requiring Natural language understanding (NLU), Natural language generation (NLG), and Natural language reasoning (NLR)?
    \item What are the recent developments in the adoption of hybrid approaches for NLU, NLG, and NLR tasks?
    \item What kinds of symbolic representations and corresponding resources are used in the literature for the development of hybrid solutions in NLP?
    \item What are the challenges faced during the adoption of hybrid solutions to the field of NLP?
    \item What are the immediate future research directions involving hybrid NLP?
\end{itemize}

\begin{figure}
\centering
\includegraphics[width=13.7cm]{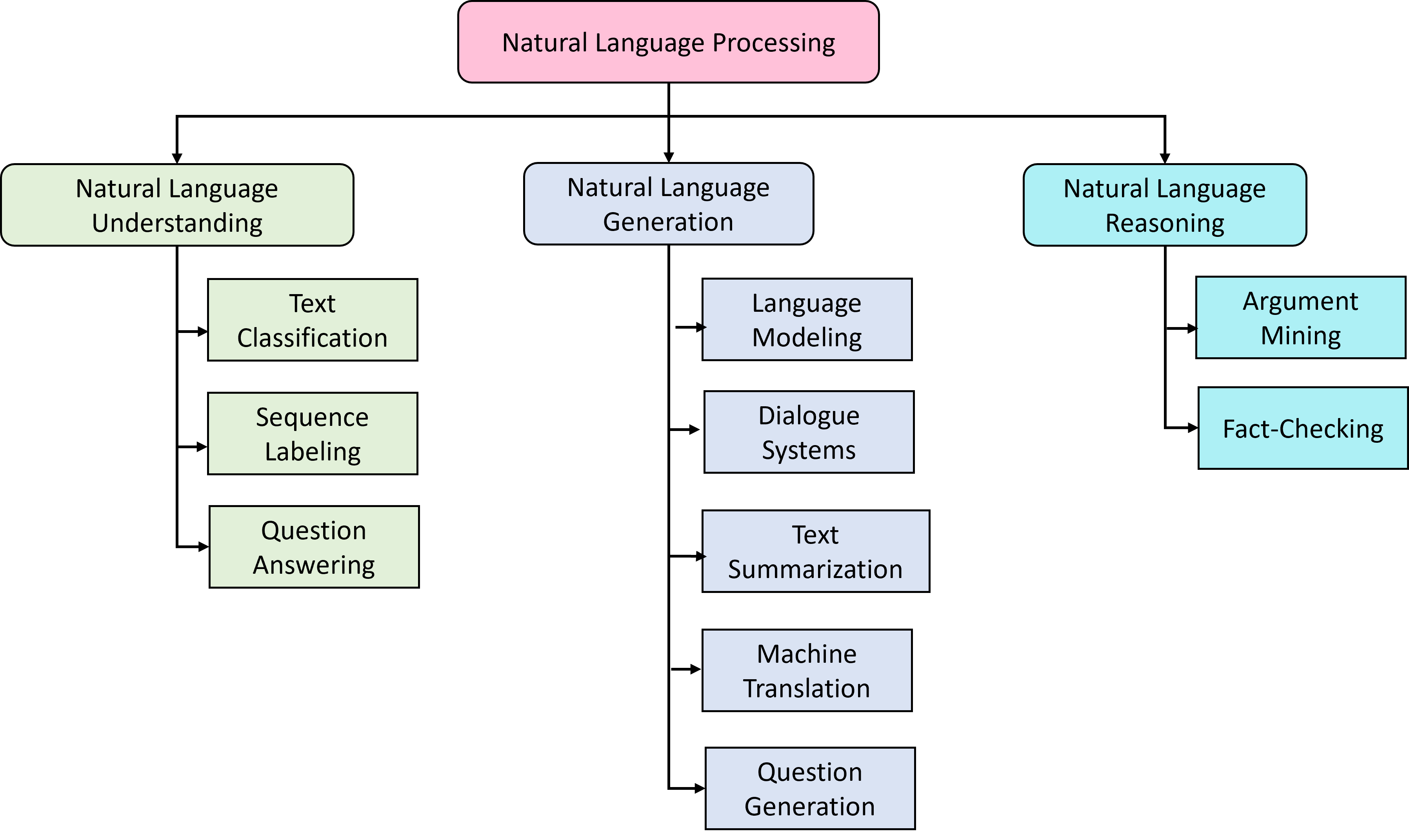}
\caption{Taxonomy of Natural Language Processing Applications}
\label{fig:nlp-applications}
\end{figure}

Figure \ref{fig:nlp-applications} depicts the NLP applications discussed in this survey. 

The closest study of our survey paper was presented by \citet{zhu2023knowledge} during a recent tutorial. The authors briefly introduced the knowledge augmentation methods used for NLP. Apart from this study, many survey articles have focused on a particular aspect of machine learning (e.g. large language models \citep{safavi2021relational,hu2023survey,yin2022survey,pan2023unifying}) or symbolic methods (e.g. knowledge graphs \citep{schneider-etal-2022-decade}) or a particular NLP task (e.g. text generation \citep{yu2022survey}) employing hybrid solutions. Further details on the related surveys are discussed in Section \ref{sec:further-reading}. To the best of our knowledge, none of the existing studies present the hybrid approaches used for NLP tasks in a wider outlook. Specifically, we discuss state-of-the-art hybridization techniques combining machine learning and symbolic methods for a wide range of NLP tasks requiring, NLU, NLG, and NLR. For each NLP task, we analyzed the research articles retrieved using the keywords \textit{Symbolic methods}, \textit{Neuro-symbolic}, \textit{Knowledge-augmented}, \textit{Knowledge-enriched}, \textit{Knowledge-aware}, and \textit{Knowledge base} along with the task name in Google Scholar, as well as other articles related to or citing those. After an in-depth analysis of research articles published during the past 6 years (2018 - 2023), we chose the state-of-the-art hybrid approaches used for each NLP task, and present them in this survey paper. Articles were selected for inclusion in the survey based on relevance to the use of hybrid approaches. 

\begin{figure}
\centering
\includegraphics[width=9cm]{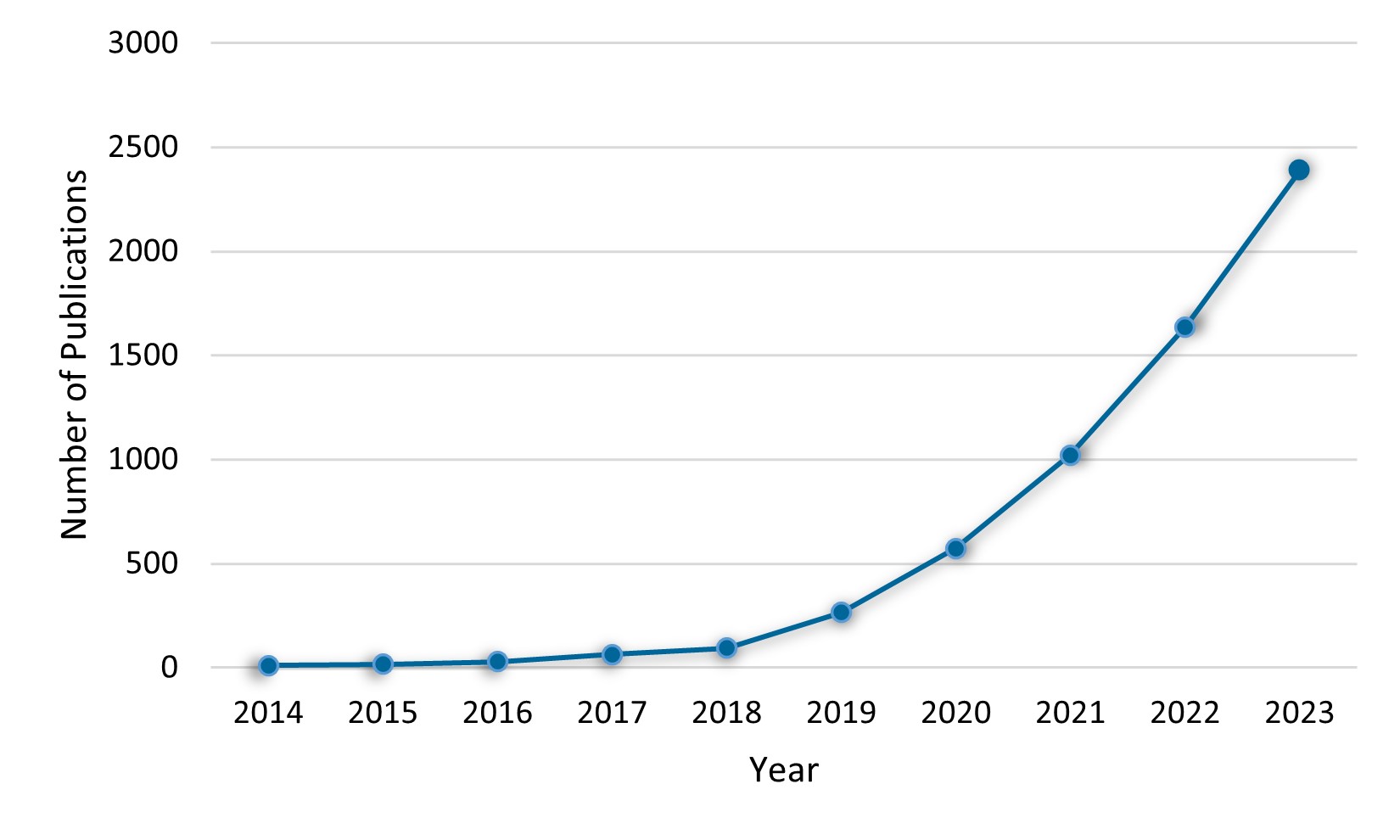}
\caption{Number of Papers Published related to Hybrid NLP}
\label{fig:publication trend}
\end{figure}

Figure \ref{fig:publication trend} shows the trend on the number of publications retrieved using the keywords \textit{Knowledge-augmented NLP, Knowledge-enriched NLP, Knowledge-aware NLP, Neuro-symbolic NLP}\footnote{https://www.dimensions.ai/}. This shows the surge of interest in the topic of \textit{Hybrid NLP}, which in turn motivates our survey detailing the current research trend in this topic and the potential future directions.

The remainder of the paper is organized as follows. Section \ref{sec:background} introduces the background knowledge required to understand the rest of the paper. State-of-the-art hybrid approaches proposed in the literature are presented in Section \ref{sec:NLU}, Section \ref{sec:NLG}, and Section \ref{sec:NLR} focusing on three fundamental objectives of Natural language processing, NLU, NLG, and NLR respectively. Section \ref{sec:challenges} discusses open challenges in the development of hybrid approaches. We discuss related survey articles as well as how they differ from ours in Section \ref{sec:further-reading}. Further, we detail future directions in Section \ref{sec:future-directions}. Finally, we conclude this review in Section \ref{sec:conclusion} along with a discussion of possible future directions. 


\section{Background}\label{sec:background}
This section introduces machine learning and symbolic methods, and the latest advancements in both fields. Further, we define the term \textit{hybrid approaches} with respect to the context of this review. 

\subsection{Machine Learning}
Machine learning has a long history of development, especially in NLP with the transformation from statistical models such as Naive Bayes classification, Hidden Markov models, Support Vector Machines, and Logistic regression to powerful neural models. Specifically, the transformers \citep{vaswani2017attention} proposed in 2017 was a breakthrough in the evolution of neural architectures. The model comprised layers of encoders and decoders with multi-headed attention, enabling them to be very effective and powerful in sequence transduction tasks. Later, the architecture was explored by other researchers to introduce several influential models, resulting in a family of transformer models. Some of the notable models include BERT (Bidirectional Encoder Representations from Transformers) \citep{bert}, GPT (Generative Pre-trained Transformer) \citep{GPT}, BART (Bidirectional Auto-Regressive Transformers) \citep{bart}, and their variations such as RoBERTa \citep{liu2019roberta}, XLM-r \citep{xlm-r}, Sentence-BERT \citep{sentence_bert}. Often these pre-trained models are fine-tuned with limited training data to achieve state-of-the-art performance in various fields related to machine learning, including NLP.

\subsection{Symbolic Methods}
\begin{figure}
\centering
    \begin{subfigure}[b]{0.75\textwidth}
    \includegraphics[width=\textwidth]{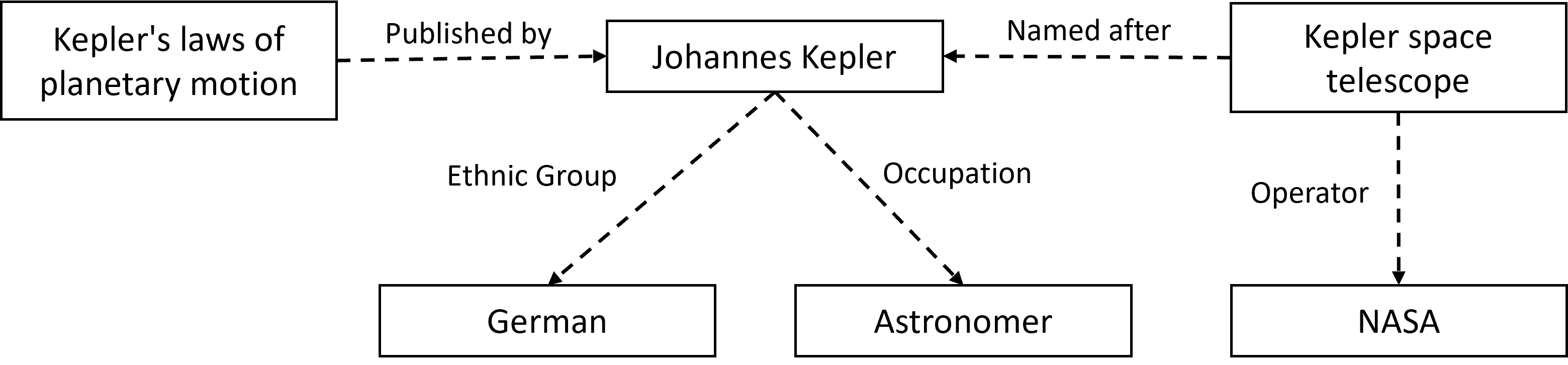}
    \caption{An Example for Knowledge Graph from Wikidata5M}
    \label{fig:KG Example}   
    \end{subfigure}  
    \begin{subfigure}[b]{0.65\textwidth}
    \includegraphics[width=\textwidth]{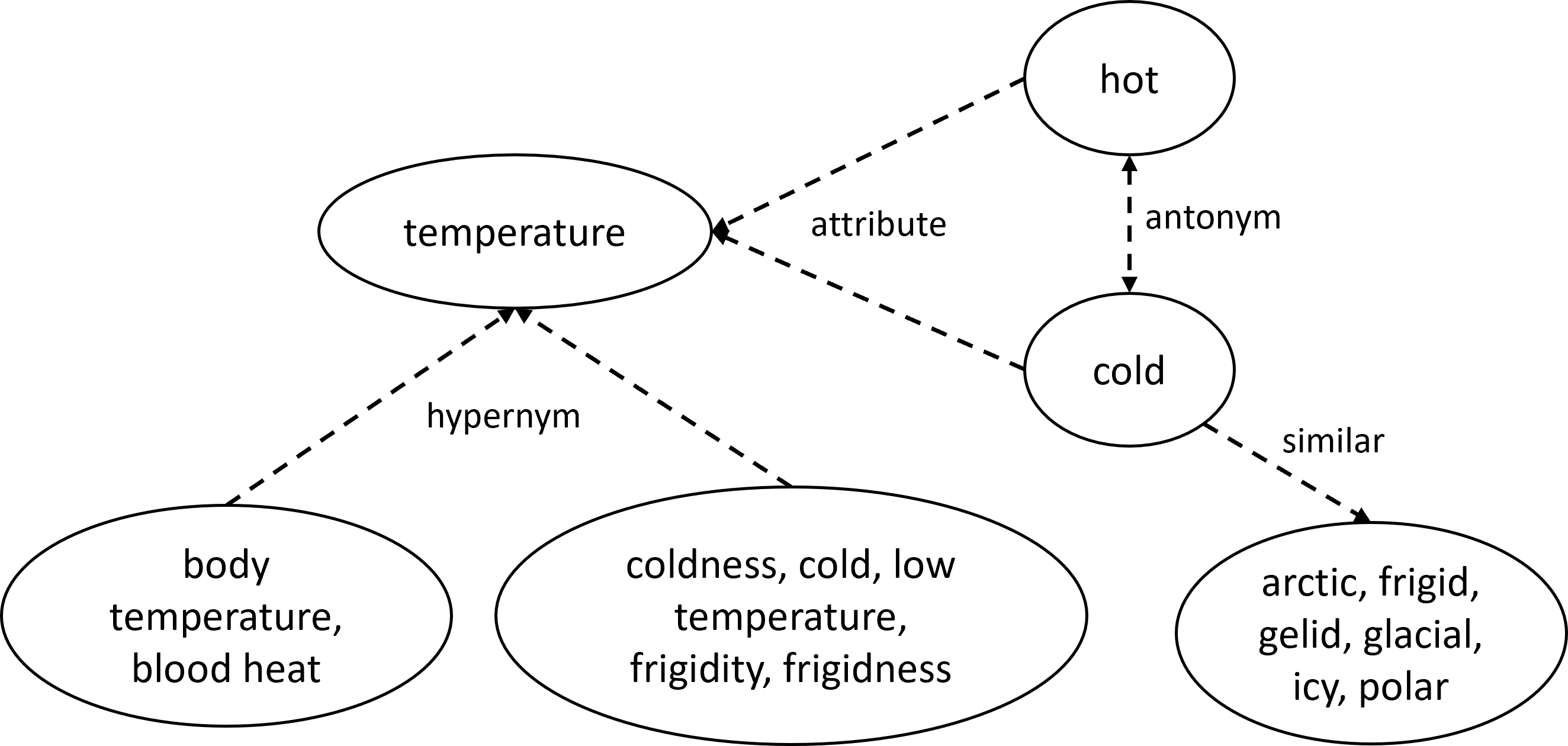}
    \caption{An Example for Ontology from WordNet}
    \label{fig:WordNet Example}   
    \end{subfigure}
    \begin{subfigure}[b]{0.72\textwidth}
    \includegraphics[width=\textwidth]{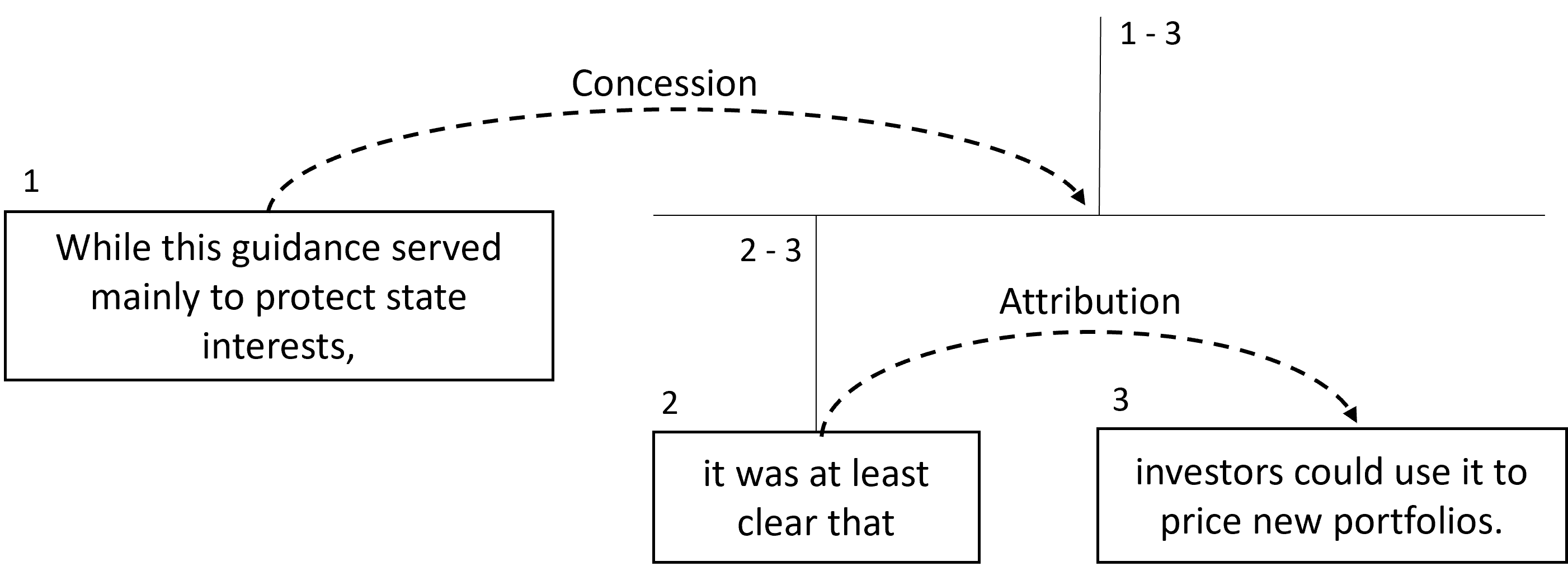}
    \caption{An Example for RST Graph}
    \label{fig:RST Example}   
    \end{subfigure}
\caption{Examples for KG, Ontology, and RST Graph.}
\label{fig2}
\end{figure}

Similar to how humans treat words and sequences of words for communication, symbolic methods simulate this cognitive behavior, by considering entities referred to as \textit{symbols} to refer to the basic unit of information in a symbolic system. The symbols can be obtained from different data structures and processed to produce new symbols \citep{hoehndorf2017data}. Numerous symbolic representations have been used to explicitly capture the knowledge. The following are some of the notable symbolic representations apart from the general graph structure.
\begin{itemize}
    \item Knowledge base (KB) - Knowledge base refers to a collection of items usually associated with an in-built index, mapping the index to items \citep{zouhar2022artefact}. The item can be either structured or unstructured conveying a piece of information in any mode. Wikipedia\footnote{https://www.wikipedia.org/} is an example of a knowledge base with a semi-structured collection of documents. Knowledge graphs and ontologies, discussed in what follows, are examples of structured knowledge bases.  
    \item Knowledge graphs (KG) - KG stores inter-related factual knowledge of entities in a directed labeled graph. The nodes represent the entities and the edges denote the relationship between the two entities. This enables the retrieval of information about entities from a knowledge graph in the form of a triplet, \textit{<head entity, relationship, tail entity>} \citep{pan2023unifying}. Figure \ref{fig:KG Example} shows an example of entities and their relationship from Wikidata5M KG.\footnote{https://deepgraphlearning.github.io/project/wikidata5m} Table \ref{tab:KG-summary} summarizes the knowledge graph types used in the literature.
    \item Ontologies - Ontologies are referred to as a formal, explicit specification of a shared conceptualization. They store concepts and their relationship (e.g. hypernyms) in a graph structure, and serve as the basis for KGs. Some of the well-known ontologies include WordNet \citep{fellbaum2010wordnet}, Probase \citep{wu2012probase}, and FrameNet \citep{baker2003structure}. Figure \ref{fig:WordNet Example} shows an example from the WordNet ontology\footnote{https://lexicala.com/review/2020/mccrae-rudnicka-bond-english-wordnet/}.  
    \item Rhetorical Structure Theory (RST) graphs - RST describes the organization of natural text and the relationship between their parts. The text structure is identified as a tree, explaining the transition point of the relations and the extent of the relation \citep{hou2020rhetorical}. The construction of the RST graph is referred to as RST parsing, which involves the identification of roles for different granularity of text (phrases, sentences, paragraphs, and collection of paragraphs). Figure \ref{fig:RST Example} shows an example of RST graph \citep{hou2020rhetorical}. 
\end{itemize}

\begin{table}
\caption{Summary of Knowledge Graph Types.}
\centering
\label{tab:KG-summary}
\scriptsize
\hspace*{-3em}
		\begin{tabular}{llp{6cm}}
			\toprule
Type               & Example                                                     & Notable Feature                                                                                                                                   \\ \midrule
\multirow{6}{*}{Encyclopedic KG}    & WikiData \citep{vrandevcic2014wikidata}  & Constructed from Wikipedia                                                                                                                        \\
                                    & Freebase   \citep{bollacker2008freebase} & Stores more structured Wikipedia data                                                                                                             \\
                                    & Wikidata5M \citep{wang2021kepler}        & Combines WikiData and Wikipedia via entity linking                                                                                                \\
                                    & Dbpedia   \citep{auer2007dbpedia}        & Constructed from Wikipedia and links other KBs                                                                                        \\
                                    & YAGO \citep{suchanek2007yago}            & Constructed from Wikipedia, GeoNames, and WordNet                                                                                                 \\
                                    & NELL \citep{carlson2010toward}           & Derived from web content                                                                                                                          \\ \midrule
\multirow{6}{*}{Commonsense KG}     & ConceptNet   \citep{speer2017conceptnet} & Covers a wide range of commonsense concepts                                                                                                       \\
                                    & ATOMIC \citep{sap2019atomic}             & Covers everyday commonsense inferential knowledge                                                                                                 \\
                                    & ATOMIC$^{20}_{20}$   \citep{hwang2021comet}     & Covering social, physical, and      eventive aspects of everyday inferential knowledge                \\
                                    & ASER \citep{zhang2020aser}               & An eventuality KG, with events as nodes and discourse relations as edges \\
                                    & TransOMCS   \citep{zhang2020transomcs}   & Derived from ConceptNet and ASER                                                                                                                  \\
                                    & CausalBanK \citep{li2021guided}          & Covers commensense related to causal                                                                                                                 \\ \midrule
\multirow{4}{*}{Domain-specific KG} & UMLS \citep{bodenreider2004unified}      & Specific to medical domain                                                                                                                        \\
                                    & Finance KG \citep{liu2019anticipating}              & Specific to finance domain                                                                                                                       \\
                                    & Geo KG \citep{zhu2017intelligent}               & Specific to geology domain  \\                           
                                    & BKG \cite{HUO2022102980} & Bibliography KG \\ \midrule
\multirow{3}{*}{Multimodal KG}      & IMGpedia \citep{ferrada2017imgpedia}     & Incorporates both text and image of DBpedia resources                                                                                             \\
                                    & MMKG \citep{liu2019mmkg}                 & Incorporates both text and image of Freebase entities                                                                                             \\
                                    & Richpedia   \citep{wang2020richpedia}    & Incorporates both text and image of Wikepedia entities \\ \bottomrule                                                                                           
\end{tabular}
\end{table}

\subsection{Hybrid Approaches}
The term \textit{hybrid} is used often in the scientific domain referring to the synergization of two aspects for solving a problem. Especially the following scenarios are considered as \textit{hybrid} in computer science research.
\begin{itemize}
    \item Hybrid models - Combining different types of models or architectures to solve a problem, e.g. integrating statistical machine learning approaches with deep learning, combining two types of deep learning models such as Long short-term memory (LSTM) network and Convolutional Neural Network (CNN). Hybrid models are often used to overcome the limitations of individual models via integration.
    \item Hybrid systems - Combining multiple solutions to solve a single goal, e.g. classifying items and clustering the items classified as relevant for the primary task. Hybrid systems are often employed to achieve multiple objectives in pursuit of a primary goal. 
    \item Hybrid data - Combining different types of data (e.g. different data structures, different modalities, different data sources) to achieve a common goal. Hybridizing data gives access to rich sources of information to solve the problem. However, the challenge in handling hybrid data escalates and requires more techniques to retrieve and process the hybrid information.    
\end{itemize}

\begin{figure}
\centering
\hspace*{-5em}
\includegraphics[width=17cm]{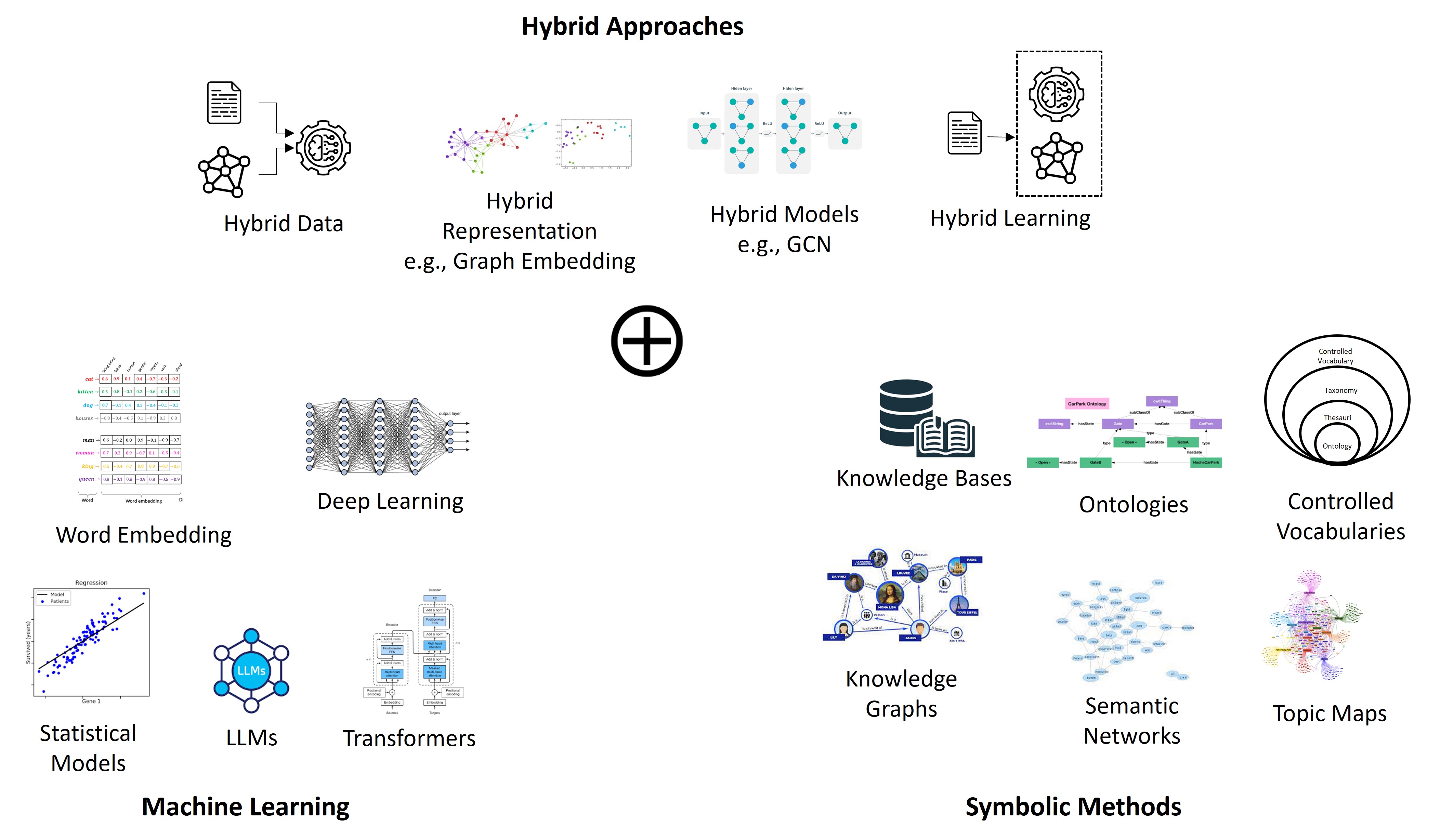}
\caption{Summary of Machine Learning Approaches, Symbolic Methods, and Hybrid Approaches}
\label{fig:hybrid-nlp}
\end{figure}

In this survey, we focus on the synergization of machine learning and symbolic methods for natural language processing as \textit{hybrid} approach, and the hybridization of these two techniques may lead to hybrid models or systems or data. Figure \ref{fig:hybrid-nlp} summarizes the machine learning approaches, symbolic methods, and hybrid approaches discussed in this paper.

Compared to the other two hybridization settings, hybrid data is a widely used technique for injecting symbolic representation into machine learning models. The knowledge bases can be directly queried using searching techniques such as Elastic search\footnote{https://www.elastic.co/} or can be linked using techniques such as entity linking \citep{schneider-etal-2022-decade} and keyword matching. The retrieved information can either be directly injected into machine learning models or transformed into vector representations. The pre-trained language models and topic models can be used to obtain vector representation for textual information retrieved from knowledge bases. Similarly, graph embedding techniques can be used to generate vector representations for nodes and edges retrieved. Table \ref{tab:emebedding-summary} presents the graph embedding techniques used in the literature. 

\begin{table}
\caption{Graph Embedding Techniques.}
\centering
\footnotesize
\label{tab:emebedding-summary}
	\begin{tabular}{lp{6cm}l}
			\toprule
Type                         & Example                                                                                                                                                     & Applicable Knowledge Base      \\ \midrule
\multirow{2}{*}{Tensor factorization methods} & RotatE   \citep{sun2019rotate} & \multirow{2}{*}{KG}                             \\
& QuantE \citep{zhang2019quaternion}          &                              \\ 
\midrule
\multirow{2}{*}{Translation-based methods}    & TransE   \citep{bordes2013translating} & \multirow{2}{*}{KG}                              \\
& TransA \citep{jia2016locally}       &                              \\ 
\midrule
\multirow{2}{*}{Neural Network-based Methods} & Graph Convolutional Networks (GCN)   \citep{schlichtkrull2018modeling} & \multirow{2}{*}{Graphs}                          \\ 
&  Graph Attention Networks (GAN) \citep{velickovic2017graph} &                          \\\midrule
\multirow{2}{*}{Graph Traversal Methods}      & Node2Vec   \citep{grover2016node2vec} & \multirow{2}{*}{Graphs}                          \\ 
&  Struc2Vec \citep{ribeiro2017struc2vec} &                          \\
\midrule
Language Modeling            & WordNet Embedding   \citep{saedi-etal-2018-wordnet}                                                                                                   & KB with associated text corpus \\ \bottomrule
\end{tabular}
\end{table}


\section{Hybrid Approaches to Natural Language Understanding}\label{sec:NLU}
Natural Language Understanding (NLU) is a branch of Natural Language Processing (NLP) that focuses on understanding the meaning and semantics of text. This includes various downstream NLU tasks such as text classification, sequence labeling, and question answering. This section briefly introduces the latest hybrid approaches used to produce state-of-the-art performance in popular NLU applications. 

\subsection{Text Classification}\label{text-classification}
Text classification is the task of automatically assigning a label from a predefined set of labels for an input text \citep{sebastiani2002machine}. While there are enormous amounts of text classification tasks explored across various domains, sentiment analysis, stance detection, and language detection are some of the prominent tasks that attracted considerable focus of the NLU research community. With the introduction of large language models and their exceptional performance across various NLP tasks, text classification tasks generally exploit fine-tuning language models on limited domain-specific training data \citep{pittaras2023content}.

A straightforward approach for adopting hybrid approaches for text classification is by incorporating external knowledge obtained via symbolic methods as additional input to the classification models.
\citet{vskrlj2021tax2vec} used word taxonomies from WordNet \citep{fellbaum2010wordnet} to generate new semantic features for text classification. The authors transformed the input document into a semantic feature representation by extracting hypernyms of words from the documents and obtaining their double normalized TF-IDF (Term Frequency - Inverse Document Frequency) scores, followed by a wide range of feature selection techniques. The authors observed a significant improvement in the performance across six short text classification tasks when these external features were used with neural classifiers. Similarly,
\citet{liu2022combining} extracted concept words from the Probase knowledge base \citep{wang2010probase} and obtained a concept word embedding by aggregating word embeddings of all the concept words present in a text. This knowledge-enriched vector representation was provided as additional input to a neural model for text classification. 

Instead of constructing a knowledge-rich vector representation as additional input,
\citet{liu2020k} generated a sentence tree as input to the BERT model by injecting knowledge graph triplets to corresponding places in the sentence. Injected knowledge was controlled via techniques such as soft-positioning and visible matrix. The authors experimented with three language-specific knowledge graphs and demonstrated promising results for the knowledge-injected BERT named K-BERT in 12 NLP tasks including text classification, sequence labeling, and question answering.
\citet{lidomain} followed the same approach and showed that developing a domain-specific knowledge graph benefits more when the knowledge graph triplets are injected into the input text.

An alternative to injecting external knowledge as inputs to classifiers is using \textit{hybrid architectures} which can combine symbolic representation learning models with traditional classifiers to improve the inference. One of the pioneering works in this direction for text classification was experimented by
\citet{yao2019graph}. The authors proposed TextGCN, a text graph convolutional network neural network architecture that models documents and words as nodes in a graph to generate a heterogeneous graph. Embedding representation for words and documents or nodes in the graph is jointly learned as supervised learning while performing text classification. This enabled the authors to transform the text classification problem into a node classification problem in a heterogeneous graph. TextGCN was shown to outperform standalone neural models in several text classification tasks. Extending TextGCN, various graph convolution network architectures \citep{gu2023enhancing} were proposed in the literature for text classification.   

\subsection{Sequence Labeling}\label{sequence labeling}
Sequence labeling is the task of assigning a label at the word level instead of the sentence or phrase level from a predefined set of labels \citep{el2022adasl}. Named entity recognition, part-of-speech tagging, and language detection are some of the well-known sequence labeling tasks. Information predicted at a word level in sequence labeling tasks generally serves as input features for various other downstream NLP tasks.

Similar to injecting external knowledge extracted from a knowledge graph into an input text, several attempts have been made to adopt similar hybrid techniques for sequence labeling tasks. In particular, the entities present in a text can be linked with external information for knowledge-augmented learning. Motivated by that direction, Enhanced Language RepresentatioN with Informative Entities (ERNIE) \citep{zhang-etal-2019-ernie} was proposed by \citeauthor{zhang-etal-2019-ernie}
to learn language representation with infused entity knowledge. The authors identified entities in the text first, aligned them with a knowledge graph, and obtained their entity embeddings using the graph embedding technique TransE \citep{bordes2013translating}. Following that, the authors trained an auto-encoder architecture with random entities masked in the Wikipedia text corpus to enforce the representation learning to incorporate entity knowledge. This model was shown to outperform BERT-based architectures in various NLP tasks including sequence labeling with limited finetuning \citep{wang2020ernie}. The underlying idea of ERNIE was later extended by replacing the auto-encoder with BERT-base architectures for sequence labeling \citep{hu2022kgner}.

Linking text with Wikipedia data for improving sequence labeling, especially for recognizing named entities, has been explored widely in recent research works \citep{tedeschi2021wikineural, wang-etal-2022-damo, boros2022knowledge}.
\citet{tedeschi2021wikineural} exploited Wikipedia data to automatically create annotated training data. The authors utilized one-to-one linkage between Wikipedia articles to generate named entity candidates and automatically annotated them as abstract concepts or named entities using the knowledge bases WordNet \citep{fellbaum2010wordnet} and BabelNet.\footnote{https://babelnet.org/} Different from this approach,
\citet{wang-etal-2022-damo} used Wikipedia as an external knowledge source to extract information related to the input sentence. The authors used Elastic search to extract relevant content from the Wikipedia articles by considering the input sentence as a query. The retrieved contexts were injected into the XLM-R \citep{xlm-r} model along with the input text for sequence classification. 

\citet{boros2022knowledge} experimented with the injection of both the relevant context retrieved using Elastic search on Wikipedia and knowledge graph embedding for knowledge-enriched training. The authors used Wikidata5M \citep{wang2021kepler}, a large-scale knowledge graph, and the RotatE embedding model \citep{sun2019rotate} to generate the embedding of entities in the knowledge graph. Similar to the search in Wikipedia articles, an embedding-based search was performed in the knowledge graph to retrieve entity embeddings similar to the input document's embedding representation. Finally, the authors convert the relevant context into embedding representation using Sentence-BERT \citep{sentence_bert} and inject both context embedding and entity embedding into the classification model. Apart from these techniques, graph neural architectures were also explored for sequence labeling tasks by converting the word-level classification problem into a node classification problem in a graph \citep{gui2019lexicon}, enabling the authors to capture non-sequential dependencies via a graph structure.   

\subsection{Question Answering}
Question Answering (QA) is the task of generating or finding relevant answers for a question in natural language \citep{zaib2022conversational}. It is one of the notable NLU tasks that often requires commonsense and external knowledge, hence demands the adoption of hybrid solutions. While training large language models in Wikipedia data and books enabled them to show exceptional performance in QA tasks \citep{mitra2019additional}, numerous other hybrid solutions have been explored in the literature recently.       

Similar to other NLU applications, an evident hybrid technique is to provide external knowledge required to perform the QA task as input to the inference model.
\citet{mitra2019additional} performed an Elastic search using question and answers as a query in external sources and retrieved the top 50 related sentences. Authors reranked the retrieved sentences using sentence similarity and provided the top 10 sentences as additional information to a BERT model. For effective context retrieval,
\citet{karpukhin-etal-2020-dense} proposed obtaining dense vector representation of passages using both the TF-IDF approach and a BERT model. During the inference, the dot-product between the dense representation of the question and the passage was used to determine the relevant context to be used as the external input to the QA model. Similarly, \citet{NORASET2021102431} experimented with different document retrieval mechanisms to extract information from Wikipedia. The authors used the BM25F retrieval algorithm \citep{perez2010using}, Google search API, and TF-IDF for retrieving the relevant content, and injected it as additional information to a Bi-LSTM model. The experiment results showed the content retrieved using the BM25F algorithm was very effective for QA.

The integration of knowledge graphs into QA tasks is an expected development in the progression of hybrid approaches.
\citet{lv2020graph} extracted evidence from the structured knowledge base ConceptNet \citep{speer2017conceptnet} and plain texts of Wikipedia articles and generated knowledge graphs for both sources for further inference. A graph convolution network (GNN) was used to extract node representation from the two knowledge graphs and the obtained contextual representation of questions and answers was fed into another graph-based attention model to choose the right answer. In addition to the enriched information available in each node in the knowledge graph,
\citet{feng-etal-2020-scalable} proposed techniques to utilize relational paths using a Multi-hop graph relation network, leading to more interpretable models. The proposed approach chooses a sub-graph related to the input query and both node embedding and path embeddings are obtained using a GNN architecture. Finally, the correct answer was chosen using a text encoder model given the embedding representations of questions and answers.
\citet{yasunaga-etal-2021-qa} extended this work to obtain a subgraph related to the QA context and then score each node in the subgraph using a language model. Later the authors jointly learned the representation of QA context and nodes in the graph using a graph neural network. The proposed model was shown to outperform other knowledge graph augmented language models in QA tasks in commonsense and biomedical domains. Following this approach, similar ideas of mutually exchanging information between language models and knowledge graphs were also explored in the literature for the QA task \citep{zhang2022greaselm,ZHANG2023103297}.   

\begin{table}
\caption{Summary of Hybrid Techniques used for NLU Tasks.}
\centering
 \scriptsize
\label{tab:NLU}
		\begin{tabular}{lp{4cm}cccccc}
			\toprule
\multicolumn{1}{l}{}                   &                                                                    & \multicolumn{5}{c}{Hybrid Approach}                                                                                                                                                                                                                         \\ \midrule
\multirow{2}{*}{Task}                  & \multicolumn{1}{c}{\multirow{2}{*}{Research}}                      & \multicolumn{3}{c}{Input}                                                                          & \multicolumn{1}{c}{\multirow{2}{*}{Model}} & \multicolumn{1}{l}{\multirow{2}{*}{\begin{tabular}[l]{@{}c@{}}Transfer \\Learning\end{tabular}}} & \multicolumn{1}{l}{\multirow{2}{*}{\begin{tabular}[l]{@{}l@{}}Training Data \\      Construction\end{tabular}}} \\ \cmidrule{3-5}
                                       & \multicolumn{1}{c}{}                                               & \multicolumn{1}{c}{\begin{tabular}[l]{@{}c@{}}KB\\Text\end{tabular}} & \multicolumn{1}{c}{\begin{tabular}[l]{@{}c@{}}KG\\Triplet\end{tabular}} & \multicolumn{1}{c}{\begin{tabular}[l]{@{}c@{}}KG\\Embedding\end{tabular}} & \multicolumn{1}{c}{} & \multicolumn{1}{c}{}                       & \multicolumn{1}{c}{}                                                                                      \\ \midrule
\multirow{3}{*}{Text   Classification} & \citet{vskrlj2021tax2vec,liu2022combining}  & \checkmark   & -  & -    & -                         & - & -         \\
           & \citet{liu2020k,lidomain}                      & - & \checkmark   & -    & -                         & - & -         \\
           & \citet{yao2019graph,gu2023enhancing}           & -  & -  & -   & \checkmark                          & - & -         \\ \midrule
\multirow{6}{*}{Sequence Labeling}     & \citet{wang2020ernie}                           & -  & - & \checkmark     & -                         & - & -         \\
           & \citet{hu2022kgner}                             & - & \checkmark   & -    & -                         & - & -         \\
           & \citet{tedeschi2021wikineural}                  & -  & -  & -    & -                        & - & \checkmark          \\
           & \citet{wang-etal-2022-damo}                    & \checkmark   & -  & -    & -                         & - & -         \\
           & \citet{boros2022knowledge}                     & \checkmark   & - & \checkmark     & -                         & - & -         \\
           & \citet{gui2019lexicon}                          & -  & -  & -   & \checkmark                          & - & -         \\ \midrule
\multirow{4}{*}{Question Answering}    & \citet{mitra2019additional,karpukhin-etal-2020-dense,NORASET2021102431}                   & \checkmark   & -  & -    & -                         & - & -         \\
           & \citet{lv2020graph}                            & \checkmark   & - & \checkmark    & \checkmark                          & - & -         \\
           & \citet{feng-etal-2020-scalable}  & -  & - & \checkmark    & \checkmark                          & - & - \\
           & \citet{yasunaga-etal-2021-qa,zhang2022greaselm,ZHANG2023103297}  & -  & - & \checkmark    & \checkmark                          & \checkmark & - \\\bottomrule        
\end{tabular}
\end{table}

Table \ref{tab:NLU} summarizes the hybrid techniques used for natural language understanding tasks. Here, we consider hybrid architectures such as Graph Neural Network modeling both distributed representation and symbolic representation as hybrid \textit{Models}. Hybrid solutions, which jointly learn the distributed representation and symbolic representation by mutually exchanging the knowledge are marked as \textit{Transfer Learning} approaches. It can be observed that the injection of external knowledge as text data and the injection of knowledge graph triplets and knowledge graph embedding are widely used as hybrid solutions across all three tasks. Further, the learning of knowledge graph embedding using graph neural networks is commonly used for question-answering tasks, possibly due to the demand of the task for retrieving supporting knowledge-enriched evidence to find the right answer. 

\section{Hybrid Approaches to Natural Language Generation}\label{sec:NLG}
Natural Language Generation (NLG) is a fundamental task in Natural Language Processing, whose aim is to produce meaningful text in natural language, generally by giving it a prompt as input. This requires semantic and syntactic understanding of the language to generate text which makes it challenging while also ensuring its applicability across a broad spectrum of NLG tasks such as dialogue systems, summarization, machine translation, and question answering. NLG solutions generally follow the encoder-decoder architecture  \citep{vaswani2017attention}, where the encoder understands the input text or prompt, and generates hidden states interpreted by the decoder to generate meaningful text \citep{yu2022survey}. However, the text generated by those powerful models often fails to match human responses due to the limited knowledge available in the training data and the lack of generalization capabilities. This demands rapid embracement of hybrid techniques to generate knowledge-enhanced text. This section presents state-of-the-art hybrid approaches used across a range of prominent NLG tasks. 

\subsection{Language Modeling}
Language modeling is the task of learning a universal representation of the language from an unlabelled text corpus \citep{rosenfeld2000two}. The task is often modeled as a next word or token prediction, where the language model is trained to predict the next word given its previous or surrounding words in a piece of text. Substantial effort has been made in the direction of integrating external knowledge into language models. 

Integration of knowledge sources into the input representation of the language models is a prominent way of infusing external knowledge prior to the inference. Analyzing in this direction,
\citet{peters-etal-2019-knowledge} performed entity linking and language modeling jointly as a multitask learning. The authors used an existing deep learning solution \citep{ganea-hofmann-2017-deep} for linking entities in the text to Wikipedia pages (i.e. for wikification) and obtained their embedding representation from entity descriptions. The resulting entity embedding was injected into the language model to generate the knowledge-enhanced representation of the text, and both entity linking and language models were jointly optimized. The authors observed an increase in the ability to recall facts in the resulting model called KnowBERT. Instead of linking entities,
\citet{ji-etal-2020-language} extracted the relevant subgraph from a knowledge graph using the Multi-hop technique and input their embedding representation by aggregating the node embedding obtained via a graph neural network. This concept representation was combined with the output of the encoder for predicting the next word.
\citet{liu2021kg} extended this idea by modifying the encoder-decoder architecture with a dedicated encoder and decoder augmented with an embedding representation obtained from the knowledge graph. Further, a dedicated convolutional neural network was used to generate the vector representation for concept words from the subgraph. Following the utilization of node embeddings as input for language representation learning, jointly optimizing the node embedding representation as well as the language representation is observed as a promising direction of improvement \citep{wang2021kepler,yu2022jaket}. 

Different from these approaches,
\citet{xiong2019pretrained} proposed to modify the training objective to force the language model to learn about real-world entities. The authors identified and linked entities mentioned in the text to Wikipedia and generated negative statements of the corresponding text by randomly replacing the entity occurrences with the names of the same entity types. During the training, the model learns to identify the correct entity. Similarly,
\citet{zhou2020pre} modified the training objective to enforce the model to generate concept-aware text. The first objective imposed the model to predict the original sentence given some unordered keywords of the sentence, whereas the next objective was aimed at recovering the order of concepts in a sentence given a shuffled list of concepts. Here, the authors define verbs, nouns, and proper nouns present in a sentence as concepts. Instead of modifying the training objective,
\citet{guan2020knowledge} post-trained the language models on sentences reconstructed from a knowledge graph. The authors converted commonsense triplets from ConceptNet \citep{speer2017conceptnet} and ATOMIC \citep{sap2019atomic} into readable sentences using a template-based method \citep{levy2017zero}.   

Generating text with complex ideas may require capturing the knowledge from structured or unstructured knowledge from external sources. Recently, this objective has been studied as a Knowledge graph to text generation problem, where the information available from external sources is converted into a knowledge graph first and the output text is generated based on the knowledge graph. In this direction,
\citet{koncel-kedziorski-etal-2019-text} propose GraphWriter, an extension of the transformer model for knowledge-graph-to-text generation. Following this study, various extensions of it were proposed \citep{cai2020graph} for encoding structural information in the knowledge graph.   

\subsection{Dialogue Systems}
Dialogue systems are designed to coherently converse with humans in natural language \citep{ni2023recent}. This requires understanding the language, recalling the conversation history, and producing accurate responses. Undoubtedly, the ability to generate precise responses hinges on the understanding of the external world and utilizing commonsense.

One of the pioneering works in the hybrid application for dialogue systems was experimented by
\citet{ghazvininejad2018knowledge}. The authors introduced two encoders dedicated to encoding the conversation history as well as the external facts, enabling the responses to be conditioned on both factors. The authors extracted focus phrases containing entities from the input query and collected raw text related to the focus phrases from external sources such as Wikipedia using entity linking techniques. A Recurrent Neural Network (RNN) encoder is used as a fact-encoder to convert the raw text with related facts into a hidden state in the proposed encoder-decoder model.
\citet{meng2020refnet} experimented with a similar solution by backing up the conversion using a domain-specific knowledge base and applying a dedicated RNN to encode the background knowledge. Instead of encoding all the related facts retrieved,
\citet{dinan2018wizard} used an attention-based component to carefully choose the relevant information gathered from external sources, and used a shared encoder to encode the knowledge and the dialogue context. Here, the TF-IDF vector representation of the articles and input query was used to retrieve relevant context from Wikipedia. The authors proposed a generative transformer memory network capable of retrieving relevant information from large memory and generating responses conditioned on both relevant information and dialogue history. 

Integration of knowledge graphs into encoder-decoder models is an anticipated research trajectory in dialogue systems study. As evidence of this,
\citet{zhou2018commonsense} proposed an encoder-decoder model coupled with graph attention mechanisms. The authors retrieved one knowledge graph per word present in the input query and converted it into a vector representation using the graph attention mechanism. This vector representation was concatenated with the vector presentation of the corresponding word and provided as input to the encoder-decoder model. The decoder model was also combined with a dynamic graph attention mechanism to attend to all the relevant knowledge graphs retrieved to generate the output. Instead of utilizing the existing knowledge graphs,
\citet{zhang-etal-2020-grounded} generated a concept graph by starting with grounded concepts present in the input and expanding it to more meaningful conversations by traversing through the related concepts. Following the other knowledge graph integration approaches, the author encoded the concept graph into a vector representation using a graph neural network and inputted to the encoder-decoder model along with the input query.  

\subsection{Text Summarization}
Text summarization is one of the core challenges in NLG which aims to generate summaries based on sources ranging from a single document to a collection of documents \citep{el2021automatic}. There are two types of underlying approaches for text summarization namely, extractive summarization and abstractive summarization. The first approach strives to choose key sentences or phrases from the source, and it is often solved as a ranking or scoring task of existing sentences in the source. On the other hand, abstractive summarization aims at producing the summary by constructing sentences or phrases using words available in the source which is commonly modeled as an NLG problem. The latest studies have shown that the summary generated by NLG models suffers from factual inconsistency issues, demanding more robust solutions.   

Aiming at resolving the factual inconsistency issue,
\citet{cao2018faithful} extracted fact descriptions from source sentences in the form of \textit{(subject, predicate, object)} using Open Information Extraction (OpenIE) tool \citep{etzioni2008open} and a dependency parser. The fact descriptions were provided as additional input to a neural model composed of two encoders and a dual attention decoder.
\citet{li-etal-2018-ensure} experimented with a similar strategy by introducing a shared encoder for external knowledge and input source, followed by a single decoder to generate the summary. A similar approach was carried out by
\citet{Wang2022KATSumKA}, where the authors extracted knowledge graph triplets related to the input text, mapped them into a low dimensional vector space and trained a graph embedding classifier to determine whether the triplet should be included in the summary or not. The embedding of triplets classified as key information was fed into a decoder along with the output of the input encoder for the summary generation. 

Topic models are also integrated with summarization models to genre topic-aware summaries in the literature. Researching in this direction,
\citet{narayan-etal-2018-dont} attempted to enforce the generation of topic-aware summaries by integrating the topic models with neural approaches. The authors first applied the topic model to the source document and input the topic distribution as an additional input of an attention-based convolutional encoder-decoder model. This enabled the model to associate each word in the document with key topics and condition the output words on the topic distribution of the document. Here, the Latent Dirichlet Allocation (LDA) model \citep{blei2003latent} was used to extract the topics from input documents. 

Similar to abstractive summarization, extractive summarization techniques have also adopted hybrid approaches to effectively model cross-sentence relations prior to the selection of summary-worthy sentences from the source.
\citet{wang-etal-2020-heterogeneous} proposed a heterogeneous graph-based neural network to model the inter-sentence relationships. The authors constructed a heterogeneous graph by modeling words and sentences as nodes in the graph. Semantic features of the nodes and edges were modeled using various techniques, including Convolutional Neural Network (CNN) and Bidirectional Long Short-Term Memory (BiLSTM) based sentence representation and TF-IDF-based edge weights. Graph attention networks (GAN) combined with transformers were used to obtain the final representation of nodes. Finally, the authors chose sentence nodes in the heterogeneous graph for summary generation via node classification. Different from this approach,
\citet{xu-etal-2020-discourse} modeled the source document as a Rhetorical Structure Theory (RST) graph and a coreference graph, to capture long-term dependencies among the discourse units in the input document. Here, the coreference graph was constructed using the entities and their coreferences. Both the document and graph were encoded using a BERT model and GCN respectively, and the encoded information was used to predict whether the input sentence should appear in the summary or not. 

\subsection{Machine Translation}
Machine translation involves the automated conversion of text from one language to another \citep{lopez2008statistical}. Initially, rule-based approaches and statistical approaches were prevalent in this field and later neural machine translation (NMT) turned out to be a key milestone in the current era. Compared to other NLG tasks, machine translation requires less information from external sources as it is enforced to preserve the content during the conversion from the source language to the target language. However, enhancing the input to NMT with linguistic features such as morphological analysis, part-of-speech tags, and dependency labels is shown to improve the quality of the task \citep{sennrich-haddow-2016-linguistic, chen2018syntax}.
\citet{bastings-etal-2017-graph} extended this idea by applying a graph convolution network on the dependency trees to obtain a dense vector representation for the sentence structure. Apart from utilizing the linguistic features,
\citet{chen2018syntax} aided the translation using search engines by extracting similar source sentences and their corresponding translation. Among the retrieved sentence pairs, top K sentences were chosen using edit distance and provided as additional input to an attention-based NMT model, enabling it to carefully attend to relevant sentence pair examples. 

\subsection{Question Generation}
The question generation task in NLG involves the automatic generation of questions from a given passage or document about a topic or context \citep{kurdi2020systematic}. This task serves as an underlying objective of various other NLG tasks such as conversation systems and plays a key role in various domains including education. While the nature of the task may vary depending on the type of question or answer \citep{mulla2023automatic}, the primary objective of question answering remains consistent: producing meaningful questions that are both syntactically and semantically accurate.

Generating questions will be centered on a certain topic or context, and the knowledge bases can serve as a rich resource of the topic or context for the question generation model. Therefore, a straightforward solution to develop a hybrid approach for question generation is either training a text generation model by extracting the text from the knowledge base as input source \citep{du-cardie-2018-harvesting} or applying rule-based techniques on structured knowledge bases such as ontologies \citep{stasaski-hearst-2017-multiple} and knowledge graphs  \citep{reddy-etal-2017-generating} to produce questions.
\citet{elsahar-etal-2018-zero} utilized both textual and structured context of the knowledge base Freebase \citep{bollacker2008freebase}. The authors extracted the triplets, text descriptions containing the subject and object of the triplet, and the phrase containing the lexicalization of the predicate of the triplet from the knowledge base. Both the textual and structured information extracted were fed as input to an attention-based encoder-decoder model for question generation. Instead of directly injecting the knowledge graph triplets as input to the question generation model,
\citet{kumar2019difficulty} observed a significant improvement in the performance, when the embedding representation of the triplet was utilized as the input. The authors used pre-trained TransE embeddings \citep{bordes2013translating} of Freebase from OpenKE tool \citep{han-etal-2018-openke} for the experiment. Following these studies, further advancements were observed in this direction by integrating graph neural networks for embedding knowledge graphs into the question generation model \citep{chen2023toward,chen2019reinforcement}. 

\begin{table}
\caption{Summary of Hybrid Techniques used for NLG Tasks.}
\centering
 \scriptsize
\label{tab:NLG}
		\begin{tabular}{lp{4cm}cccccccccc} \toprule
\multicolumn{1}{l}{}                 &                                                                                 & \multicolumn{10}{c}{Hybrid Approach}                                                                                                                                                                                                                                                                                                                                                                                                                                                                                                                                                                                                                                    \\ \midrule
\multirow{2}{*}{Task}                & \multicolumn{1}{c}{\multirow{2}{*}{Research}}                                   & \multicolumn{6}{c}{Input}                                                                                                                                                                                                                                                                                                                                                                                                            & \multicolumn{1}{c}{\multirow{2}{*}{Model}} & \multicolumn{2}{c}{Learning}                                      & \multicolumn{1}{l}{\multirow{2}{*}{\rotatebox{90}{\begin{tabular}[l]{@{}l@{}}Training Data \\Construction\end{tabular}}}} \\ \cmidrule(l){3-8} \cmidrule(l){10-11}
                                     & \multicolumn{1}{c}{}                                                            & \rotatebox{90}{KB Text} & \rotatebox{90}{KG Triplet} & \rotatebox{90}{KG Embedding} & \rotatebox{90}{Graph Emebdding} & \rotatebox{90}{KB} & \rotatebox{90}{Topic} &  & \rotatebox{90}{Transfer Learning} & \rotatebox{90}{Objective}                                                                                            \\  \midrule
\multirow{8}{*}{Language Modeling}   & \citet{peters-etal-2019-knowledge}                                                      & \checkmark                           & -& -  & -           & -& -        & -            & -   & -       & -                                                                                 \\
                                     & \citet{ji-etal-2020-language}                                                           & -     & -& \checkmark                                & -           & -& -        & \checkmark                                          & -   & -       & -                                                                                 \\
                                     & \citet{liu2021kg}                                                                       & -     & -& \checkmark                                & -           & -& -        & -            & -   & -       & -                                                                                 \\
                                     & \citet{wang2021kepler}                                                                  & -     & -& \checkmark                                & -           & -& -        & -            & \checkmark                                 & -       & -                                                                                 \\
                                     & \citet{yu2022jaket}                                                                     & -     & -& \checkmark                                & -           & -& -        & \checkmark                                          & \checkmark                                 & -       & -                                                                                 \\
                                     & \citet{xiong2019pretrained,zhou2020pre}                                                & -     & -& -  & -           & -& -        & -            & -   & \checkmark                             & -                                                                                 \\
                                     & \citet{guan2020knowledge}                                                               & -     & -& -  & -           & -& -        & -            & -   & -       & \checkmark                                                                                                               \\
                                     & \citet{koncel-kedziorski-etal-2019-text,cai2020graph}                                          & -& -  & -           & -                     & \checkmark         & -            & -   & -       & -                                                                                 \\ \midrule
\multirow{2}{*}{Dialogue Systems}   & \citet{ghazvininejad2018knowledge,dinan2018wizard,liu2019knowledge,meng2020refnet} & \checkmark                           & -& -  & -           & -& -        & -            & -   & -       & -                                                                                 \\
                                     & \citet{zhou2018commonsense,zhang-etal-2020-grounded}                                 & -     & -& \checkmark                                & -           & -& -        & \checkmark                                          & -   & -       & -                                                                                 \\ \midrule
\multirow{5}{*}{Text Summarization}  & \citet{cao2018faithful,li-etal-2018-ensure}                                            & \checkmark                           & -& -  & -           & -& -        & -            & -   & -       & -                                                                                 \\
                                     & \citet{Wang2022KATSumKA}                                                                & -     & -& \checkmark                                & -           & -& -        & \checkmark                                          & -   & -       & -                                                                                 \\
                                     & \citet{narayan-etal-2018-dont}                                                          & -     & -& -  & -           & -& \checkmark                                      & -            & -   & -       & -                                                                                 \\
                                     & \citet{wang-etal-2020-heterogeneous}                                                    & -     & -& -  & -           & -& -        & \checkmark                                          & -   & -       & -                                                                                 \\
                                     & \citet{xu-etal-2020-discourse}                                                          & -     & -& -  & \checkmark                                         & -& -        & -            & -   & -       & -                                                                                 \\ \midrule
\multirow{2}{*}{Machine Translation} & \citet{bastings-etal-2017-graph}                                                        & -     & -& -  & -           & -& -        & \checkmark                                          & -   & -       & -                                                                                 \\
                                     & \citet{chen2018syntax}                                                                  & \checkmark                           & -& -  & -           & -& -        & -            & -   & -       & -                                                                                 \\ \midrule
\multirow{5}{*}{Question Generation} & \citet{reddy-etal-2017-generating,du-cardie-2018-harvesting}                         & -     & -& -  & -           & -& -        & -            & -   & -       & \checkmark                                                                                                               \\
                                     & \citet{stasaski-hearst-2017-multiple}                                                   & -     & -& -  & -           & \checkmark                      & -        & -            & -   & -       & -                                                                                 \\
                                     & \citet{elsahar-etal-2018-zero}                                                          & \checkmark                           & \checkmark                              & -  & -           & -& -        & -            & -   & -       & -                                                                                 \\
                                     & \citet{kumar2019difficulty}                                                             & -     & -& \checkmark                                & -           & -& -        & -            & -   & -       & -                                                                                 \\
                                     & \citet{chen2023toward,chen2019reinforcement}                                                      & -& \checkmark                                & -           & -& -        & \checkmark                                          & -   & -       & -     & -  \\ \bottomrule                                                                         
\end{tabular}
\end{table}

Table \ref{tab:NLG} summarizes the existing hybrid techniques for NLG tasks. It can be observed that a wide range of techniques are adopted for NLG compared to NLU tasks, especially as hybrid inputs and learning techniques. A notable hybrid input will be the injection of the whole knowledge base as input to the machine learning model, where the requirement of the task is to convert or derive text from the knowledge base (e.g. knowledge graph to text generation, knowledge base to question generation). Hybrid learning techniques are mainly used with language modeling tasks for developing knowledge-aware and generalized models. While the injection of KB textual content, structured content, or embedding representations are widely used across many NLG tasks, text summarization tasks exploit various other hybrid approaches such as injection of RST and coreference graph embedding and topic vectors. The machine translation task demands less amount of external knowledge compared to the other NLG tasks, hence, very little attention has been given to the development of hybrid solutions for machine translation. Apart from these hybrid solutions for various NLG tasks, another promising research direction related to NLG will be the generation of knowledge-aware explanations of inferences using hybrid solutions \citep{YANG2023103245}.       


\section{Hybrid Approaches to Natural Language Reasoning}\label{sec:NLR}
Natural language reasoning (NLR) aims to integrate diverse knowledge sources such as encyclopedic and commonsense knowledge, to draw new logical conclusions about the actual or hypothetical world \citep{yu2023nature}. The knowledge can be integrated from both implicit (e.g. pre-trained language models) and explicit sources (e.g. knowledge bases). Reasoning plays a vital role in various NLU and NLG tasks, where neither memorizing the knowledge in the training data nor understanding the context is sufficient for deriving conclusions and requiring the integration of knowledge. This section presents hybrid approaches used for notable NLR tasks including argument mining, and automated fact-checking. 

\subsection{Argument Mining}
Argument mining involves the identification of structured argument data from unstructured text, including the identification of premise and conclusion \citep{lawrence-reed-2019-argument}. This enables understanding of the individual components of the arguments and their relationships used to convey the overall message. Argument mining is widely used for the development of qualitative assessment tools for online content, grabbing the attention of both policymakers and social media researchers. 

Integration of knowledge graphs, Wikipedia, search engines, and pre-trained language models are often used as a solution for argument mining to infuse external knowledge and commonsense required for the inference \citep{fromm2019tacam,abels2021focusing, saadat-yazdi-etal-2022-kevin, saadat-yazdi-etal-2023-uncovering}.
\citet{fromm2019tacam} integrated three knowledge sources (i.e. Word2Vec \citep{mikolov2013efficient}, DBpedia knowledge graph \citep{auer2007dbpedia}, and a pre-trained BERT model) for classifying a sentence as an argument or not for the given topic. The authors obtained vector representations of the sentence and the topic using the Word2Vec model and input them into a BiLSTM to encode them. Triplets of the entities present in the argument were extracted from the knowledge graph and converted into embedding vectors using a graph embedding technique TransE \citep{bordes2013translating}. Encoded topic and argument vectors and entity embeddings were used to fine-tune a BERT classifier. Similarly,
\citet{abels2021focusing} integrated topic modeling, Wikipedia, Knowledge graph, and search engine for argument mining. The authors learned topics using Latent Dirichlet Allocation (LDA) \citep{blei2003latent} from entity-linked Wikipedia pages from the input sentence. Further, they obtained the subgraph related to the topic words from the existing knowledge graph, Wikidata \citep{vrandevcic2014wikidata}. To resolve the incompleteness issue in Wikipedia, the authors constructed another knowledge graph using content that resulted in a Google search for the topic words. Finally, the authors extracted evident paths from both knowledge graphs using a breadth-first search, converted each path into a vector representation using a Bi-LSTM, and used it as additional input for argument mining.

\subsection{Automated Fact-Checking}
Automated fact-checking is an essential task for detecting and mitigating the impact of misinformation \citep{guo2022survey,zeng2021automated}. This is generally composed of three stages: 1. claim detection to identify sentences with check-worthy or verifiable claims; 2. evidence retrieval to extract supporting statements of the claim; and 3. claim verification to validate whether the retrieved claim is true or not based on the evidence. An intermediate stage of `claim matching' is sometimes added before the `evidence retrieval', where the claim matching task consists in grouping together claims that can be resolved with the same fact-check and therefore need not be treated separately in the subsequent stages \citep{kazemi2021claim}. The evidence retrieval and claim verification tasks are often combined and handled as fact verification \citep{guo2022survey}. It is evident that leveraging hybrid knowledge sources and techniques can significantly enhance fact-verification tasks for precise inference. 

Fact-checking using knowledge sources was often resolved by constructing knowledge graphs using the evidence gathered and executing path detection algorithms in knowledge graphs for claim verification  \citep{ciampaglia2015computational, shi2016fact, shiralkar2017finding} until the integration of large language models. Addressing this aspect,
\citet{zhou-etal-2019-gear} integrated Wikipedia data, the BERT model, and a graph neural network for claim verification. The authors retrieved related sentences to the claim from Wikipedia using MediaWiki API\footnote{https://www.mediawiki.org/wiki/API} and chose the top 5 relevant sentences using the hinge loss function. Both the claim and relevant sentences were encoded using the BERT model and input to a graph neural network for veracity detection of the claim.
\citet{zhong-etal-2020-reasoning} extended this approach by explicitly modeling the relationship between the evidence sentences by constructing an evidence graph. The authors applied the AllenNLP tool \citep{Gardner2017AllenNLP} for semantic role labeling and modeled arguments and links between arguments as nodes and edges in the evidence graph. A Graph-enhanced contextual representation of the words in the evidence graph was extracted by the pre-trained model XL-net \citep{yang2019xlnet} and inputted to graph neural network for veracity classification of claims. Adopting this methodology, numerous inference techniques using graph neural networks and evidence graphs for fact-checking have been employed in the literature \citep{liu-etal-2020-fine, xu2022evidence}.

Apart from the graph-based techniques,
\citet{si-etal-2021-topic} integrated topic models and neural networks for retrieving topic-constrained evidence information. Given a claim and set of evidence sentences, the authors applied Latent Dirichlet Allocation to extract the topics from the evidence sentences, and the topic distributions learned are used to obtain a topic representation of the evidence via a co-attention mechanism. This enabled the authors to incorporate topic consistency between the evidence and the topic consistency between the claim and evidence into dense representations. This topic-aware evidence representation and claim were input to a capsule network for determining the stance of the evidence towards the claim.    

It can be observed that, compared to other NLU and NLG tasks, natural language reasoning demands multiple explicit and implicit knowledge sources integrated together for deriving new conclusions. This includes knowledge bases, graph embedding solutions, pre-trained language models, and topic modeling.   


\section{Challenges}\label{sec:challenges}
While hybrid approaches eliminate weaknesses of symbolic methods and machine learning approaches, they pose certain challenges that would significantly impact practical usage. Following are some of the key challenges in the implementation of hybrid approaches.
\begin{itemize}
    \item \textbf{Generalization of knowledge:} Although existing powerful models are infused with external knowledge sources for accurate inference, machine learning models tend to remember the knowledge provided during the learning and fail to update their internal memory according to the changes in the real world. This requires more generalized solutions without the need to retrain the model with changes in the real world and adapt to temporal changes for reduced model deterioration and improved persistence \citep{alkhalifa2023building}.
    \item \textbf{Generalization across tasks:}  Hybrid models are often trained for a specific task with the integration of symbolic representation required to accomplish inference for the given task, making them incompatible across other NLP tasks. Research in model generalizability across tasks is still in its infancy, including in the development of models for zero-shot adaptation to new tasks.
    \item \textbf{Human-level reasoning:} Hybrid approaches are relatively powerful in natural language reasoning. However, human-level reasoning remains an open research problem  \citep{yin2022survey}, requiring more robust reasoning models simulating human thoughts. Especially, recent studies \citep{branco2021commonsense} have reported the inconsistent performance of hybrid models in reasoning tasks and observed that the models tend to remember the shortcuts present in the training data. Further, these models tend to rely more on reasoning on entities or the syntactic structure of the data. Hence, it is unclear whether the implicit commonsense knowledge of the hybrid models or the correlation in data results in superior performance in handling complex situations. This demands the development of more robust reasoning models encouraging the acquisition of commonsense knowledge rather than remembering training data.
    \item \textbf{Reliability of knowledge sources:} Another key concern of hybrid models is the reliability of external sources infused during the training which are often curated using automated tools and search engines. Curated data may contain biased and/or false information. Further, the knowledge related to a topic or context may vary over time, and utilizing outdated information may result in inaccurate inferences. This questions the reliability of the knowledge bases and the factual inferences obtained by hybrid models \citep{yin2022survey}.
    \item \textbf{Increased computational requirements:} Integration of KBs with machine learning models increases the requirement of both computational resources and inference power to deal with a large number of logical rules and constraints. While approximate inference can be employed as a solution at a cost of reduced performance \citep{YU2023105}, this does not serve the purpose of adopting hybrid solutions. Therefore, it is required to explore effective solutions utilizing the computational strength of neural models to tackle the increased computational requirements.
    \item \textbf{Requirement of customized KBs: } Even though most of the hybrid solutions employ existing general-purpose KBs, various studies \cite{lidomain,meng2020refnet,HUO2022102980} have shown the requirement of developing task-specific or domain-specific KBs and reported promising results when KBs are developed specifically for an NLP task or domain. This hinders the prompt adoption of hybrid solutions across other NLP tasks and domains.   
    \item \textbf{Uniform knowledge acquisition and representation:}  KBs support different retrieval systems and knowledge representation mechanisms depending on the information stored. Therefore, the existing approaches often rely on a single KB and utilize a KB-based technique to retrieve and represent the information. This restricts the utilization of multiple KBs for accomplishing an NLP task and demands more uniform approaches to retrieve and represent knowledge across various KBs \citep{zouhar2022artefact}.
\end{itemize}


\section{Related Surveys}\label{sec:further-reading}
\citet{zhu2023knowledge} briefly introduced the knowledge augmentation methods used for NLP during a recent tutorial. The study was centered around the integration of knowledge into language models for NLP. Apart from this study, various reviews have discussed the integration of knowledge graphs for NLP \citep{schneider-etal-2022-decade}, and integration of knowledge into large language models (LLMS) \citep{safavi2021relational,hu2023survey,yin2022survey,pan2023unifying}.
\citet{pan2023unifying} presented the latest survey on this direction focusing on KG-enhanced LLMs, LLM-enhance KGs, and scenarios where LLM and KG can play equal roles. Focusing on specific NLP tasks,
\citet{yu2022survey} detailed a survey on knowledge-enhanced text generation. From a more theoretical perspective,
\citet{ferrone2020symbolic} presented a survey summarizing the link between distributed and symbolic representations, and explained how symbols are represented in deep learning for NLP. Similarly,
\citet{zouhar2022artefact} systematically described artifacts (items retrieved from the knowledge base) and techniques used in the literature to retrieve and inject the artifacts into NLP models. 

\citet{hamilton2022neuro} examined the impact of neuro-symbolic approaches in NLP by analyzing the performance of the models in terms of five key criteria, out-of-distribution generalization, interpretability, reduced training data, transferability across domains, and reasoning. The authors concluded in their study, that there is no clear correlation is met between the integration of knowledge and the performance criteria analyzed.   

Different from all these studies, we focus on the broader aspect of synergizing machine learning techniques and symbolic methods for NLP and present the state-of-the-art hybrid approaches proposed in the literature for a wide range of NLP tasks, providing a unique overview of this increasingly popular trend in NLP.


\section{Future Directions}\label{sec:future-directions}
In the previous sections, we introduced the latest advancements in Hybrid NLP and the challenges linked to embracing this emerging line of research. Despite the challenges to be addressed, numerous open research problems persist in this promising research topic. We next put forth a set of immediate future research directions in hybrid NLP.
\begin{itemize}
    \item \textbf{Introduction of pre-trained hybrid models: } Training hybrid models is computationally complex, and this hinders the accessibility of hybrid solutions for researchers with diverse levels of (typically limited) training resources. Similar to the effectiveness of pre-trained language models, a comparable approach can be embraced for hybrid models. Therefore, the introduction of pre-trained hybrid models has the potential to streamline computational efficiency, making them more accessible and appealing to a broader community of researchers.
    
    \item \textbf{Zero-short or few-shot reasoning using hybrid models: } Training hybrid models requires a massive amount of quality labeled data. However, certain domains face challenges due to limited labeled data. Therefore, zero-shot or few-short learning using hybrid models can facilitate learning with limited training data. While Large Language Models (LLMs) have shown impressive performance in zero-shot/few-shot learning, this learning trend is yet to be further studied for hybrid solutions.    
    
    \item \textbf{Dynamic Reasoning: } Hybrid solutions play a key role in Natural Language Reasoning (NLR) by facilitating the generation of more meaningful and interpretable conclusions. Nevertheless, the existing hybrid solutions for NLR primarily focus on static symbolic representations. However, the information in symbolic representations may get outdated, e.g. a tuple (Donald Trump, President of, America) in a knowledge graph is no longer valid. Addressing this issue requires both the symbolic representation and reasoning to dynamically get updated. This aspect of the research remains unexplored and needs further investigation. 
    
    \item \textbf{Automatic construction of symbolic knowledge: } While the present hybrid solutions highly benefit from the existing symbolic knowledge sources, there is a growing need for knowledge sources with various characteristics, e.g. domain-specific knowledge sources. Therefore, it will be an interesting research direction to explore the possibility of automatically generating symbolic knowledge through hybrid solutions \citep{YU2023105}. 
    
    \item \textbf{Handling multimodal data}: The present focus on hybrid techniques involves the utilization of textual knowledge in knowledge bases to develop effective NLP solutions. However, much of today's real-world data is predominantly multimodal, and this necessitates the introduction of multimodal hybrid techniques to effectively harness the knowledge across different modalities. The existence of multimodal knowledge bases and the latest developments in multimodal LLMs may expedite the adoption of multimodal hybrid techniques. However, this research direction requires further investigation and advancements.
\end{itemize}

\section{Conclusion}\label{sec:conclusion}
The use of a hybrid approach is a promising direction of combining rich knowledge in symbolic methods with machine learning to enhance their inference capabilities. Moreover, it facilitates the generation of more credible and factually grounded inferences by incorporating external knowledge and commonsense, ultimately improving the reliability and adaptability of hybrid solutions across a broad spectrum of NLP tasks. Our survey paper is the first to provide a broad overview of hybrid approaches presented in three subareas of NLP including understanding (NLU), generation (NLG), and reasoning (NLR). We further delve into a set of tasks in each of these subareas, discussing state-of-the-art methods and progress made in the scientific community. We conclude the overview by discussing open research challenges for further development of hybrid approaches to NLP.

Hybrid approaches for NLP generally exploit two types of techniques, integration of symbolic knowledge into the input of statistical or deep learning models and modifying the deep learning models with symbolic structures resulting in architectures such as graph neural networks and hybrid data representation such as graph embedding. The first approach enables a wide range of adoption of symbolic knowledge sources such as external databases, knowledge graphs, and topic models and encourages the model to utilize this knowledge as additional information during inference to make accurate decisions. Similarly, hybrid architectures empower powerful representations of data and their relationships via hybrid structures, improving the scalability and explainability of the task. There is no doubt, that a clear understanding of both machine learning approaches and symbolic methods will lead to innovative architectures with substantial advancements in the future.

\section*{Declaration of competing interest}
The authors declare that they have no known competing financial interests or personal relationships that could have appeared to influence the work reported in this paper.

\section*{Acknowledgments}
This project is funded by the European Union and UK Research and Innovation under Grant No. 101073351 as part of Marie Skłodowska-Curie Actions (MSCA Hybrid Intelligence to monitor, promote, and analyze transformations in good democracy practices).



\bibliographystyle{elsarticle-harv} 
\bibliography{references,anthology}

\begin{thebibliography}{148}
\expandafter\ifx\csname natexlab\endcsname\relax\def\natexlab#1{#1}\fi
\providecommand{\url}[1]{\texttt{#1}}
\providecommand{\href}[2]{#2}
\providecommand{\path}[1]{#1}
\providecommand{\DOIprefix}{doi:}
\providecommand{\ArXivprefix}{arXiv:}
\providecommand{\URLprefix}{URL: }
\providecommand{\Pubmedprefix}{pmid:}
\providecommand{\doi}[1]{\href{http://dx.doi.org/#1}{\path{#1}}}
\providecommand{\Pubmed}[1]{\href{pmid:#1}{\path{#1}}}
\providecommand{\bibinfo}[2]{#2}
\ifx\xfnm\relax \def\xfnm[#1]{\unskip,\space#1}\fi
\bibitem[{Abels et~al.(2021)Abels, Ahmadi, Burkhardt, Schiller, Gurevych and Kramer}]{abels2021focusing}
\bibinfo{author}{Abels, P.}, \bibinfo{author}{Ahmadi, Z.}, \bibinfo{author}{Burkhardt, S.}, \bibinfo{author}{Schiller, B.}, \bibinfo{author}{Gurevych, I.}, \bibinfo{author}{Kramer, S.}, \bibinfo{year}{2021}.
\newblock \bibinfo{title}{Focusing knowledge-based graph argument mining via topic modeling}.
\newblock \bibinfo{journal}{arXiv preprint arXiv:2102.02086} .
\bibitem[{Alkhalifa et~al.(2023)Alkhalifa, Kochkina and Zubiaga}]{alkhalifa2023building}
\bibinfo{author}{Alkhalifa, R.}, \bibinfo{author}{Kochkina, E.}, \bibinfo{author}{Zubiaga, A.}, \bibinfo{year}{2023}.
\newblock \bibinfo{title}{Building for tomorrow: Assessing the temporal persistence of text classifiers}.
\newblock \bibinfo{journal}{Information Processing \& Management} \bibinfo{volume}{60}, \bibinfo{pages}{103200}.
\bibitem[{Auer et~al.(2007)Auer, Bizer, Kobilarov, Lehmann, Cyganiak and Ives}]{auer2007dbpedia}
\bibinfo{author}{Auer, S.}, \bibinfo{author}{Bizer, C.}, \bibinfo{author}{Kobilarov, G.}, \bibinfo{author}{Lehmann, J.}, \bibinfo{author}{Cyganiak, R.}, \bibinfo{author}{Ives, Z.}, \bibinfo{year}{2007}.
\newblock \bibinfo{title}{Dbpedia: A nucleus for a web of open data}, in: \bibinfo{booktitle}{international semantic web conference}, \bibinfo{organization}{Springer}. pp. \bibinfo{pages}{722--735}.
\bibitem[{Baker et~al.(2003)Baker, Fillmore and Cronin}]{baker2003structure}
\bibinfo{author}{Baker, C.F.}, \bibinfo{author}{Fillmore, C.J.}, \bibinfo{author}{Cronin, B.}, \bibinfo{year}{2003}.
\newblock \bibinfo{title}{The structure of the framenet database}.
\newblock \bibinfo{journal}{International Journal of Lexicography} \bibinfo{volume}{16}, \bibinfo{pages}{281--296}.
\bibitem[{Bastings et~al.(2017)Bastings, Titov, Aziz, Marcheggiani and Sima{'}an}]{bastings-etal-2017-graph}
\bibinfo{author}{Bastings, J.}, \bibinfo{author}{Titov, I.}, \bibinfo{author}{Aziz, W.}, \bibinfo{author}{Marcheggiani, D.}, \bibinfo{author}{Sima{'}an, K.}, \bibinfo{year}{2017}.
\newblock \bibinfo{title}{Graph convolutional encoders for syntax-aware neural machine translation}, in: \bibinfo{booktitle}{Proceedings of the 2017 Conference on Empirical Methods in Natural Language Processing}, \bibinfo{publisher}{Association for Computational Linguistics}, \bibinfo{address}{Copenhagen, Denmark}. pp. \bibinfo{pages}{1957--1967}.
\newblock \URLprefix \url{https://aclanthology.org/D17-1209}, \DOIprefix\doi{10.18653/v1/D17-1209}.
\bibitem[{Blei et~al.(2003)Blei, Ng and Jordan}]{blei2003latent}
\bibinfo{author}{Blei, D.M.}, \bibinfo{author}{Ng, A.Y.}, \bibinfo{author}{Jordan, M.I.}, \bibinfo{year}{2003}.
\newblock \bibinfo{title}{Latent dirichlet allocation}.
\newblock \bibinfo{journal}{Journal of machine Learning research} \bibinfo{volume}{3}, \bibinfo{pages}{993--1022}.
\bibitem[{Bodenreider(2004)}]{bodenreider2004unified}
\bibinfo{author}{Bodenreider, O.}, \bibinfo{year}{2004}.
\newblock \bibinfo{title}{The unified medical language system (umls): integrating biomedical terminology}.
\newblock \bibinfo{journal}{Nucleic acids research} \bibinfo{volume}{32}, \bibinfo{pages}{D267--D270}.
\bibitem[{Bollacker et~al.(2008)Bollacker, Evans, Paritosh, Sturge and Taylor}]{bollacker2008freebase}
\bibinfo{author}{Bollacker, K.}, \bibinfo{author}{Evans, C.}, \bibinfo{author}{Paritosh, P.}, \bibinfo{author}{Sturge, T.}, \bibinfo{author}{Taylor, J.}, \bibinfo{year}{2008}.
\newblock \bibinfo{title}{Freebase: a collaboratively created graph database for structuring human knowledge}, in: \bibinfo{booktitle}{Proceedings of the 2008 ACM SIGMOD international conference on Management of data}, pp. \bibinfo{pages}{1247--1250}.
\bibitem[{Bordes et~al.(2013)Bordes, Usunier, Garcia-Duran, Weston and Yakhnenko}]{bordes2013translating}
\bibinfo{author}{Bordes, A.}, \bibinfo{author}{Usunier, N.}, \bibinfo{author}{Garcia-Duran, A.}, \bibinfo{author}{Weston, J.}, \bibinfo{author}{Yakhnenko, O.}, \bibinfo{year}{2013}.
\newblock \bibinfo{title}{Translating embeddings for modeling multi-relational data}.
\newblock \bibinfo{journal}{Advances in neural information processing systems} \bibinfo{volume}{26}.
\bibitem[{Boros et~al.(2022)Boros, Gonz{\'a}lez-Gallardo, Giamphy, Hamdi, Moreno and Doucet}]{boros2022knowledge}
\bibinfo{author}{Boros, E.}, \bibinfo{author}{Gonz{\'a}lez-Gallardo, C.E.}, \bibinfo{author}{Giamphy, E.}, \bibinfo{author}{Hamdi, A.}, \bibinfo{author}{Moreno, J.G.}, \bibinfo{author}{Doucet, A.}, \bibinfo{year}{2022}.
\newblock \bibinfo{title}{Knowledge-based contexts for historical named entity recognition \& linking}.
\newblock \bibinfo{journal}{Experimental IR Meets Multilinguality, Multimodality, and Interaction} .
\bibitem[{Branco et~al.(2021)Branco, Branco, Silva and Rodrigues}]{branco2021commonsense}
\bibinfo{author}{Branco, R.}, \bibinfo{author}{Branco, A.}, \bibinfo{author}{Silva, J.M.}, \bibinfo{author}{Rodrigues, J.}, \bibinfo{year}{2021}.
\newblock \bibinfo{title}{Commonsense reasoning: how do neuro-symbolic and neuro-only approaches compare?}, in: \bibinfo{booktitle}{CIKM Workshops}.
\bibitem[{Cai and Lam(2020)}]{cai2020graph}
\bibinfo{author}{Cai, D.}, \bibinfo{author}{Lam, W.}, \bibinfo{year}{2020}.
\newblock \bibinfo{title}{Graph transformer for graph-to-sequence learning}, in: \bibinfo{booktitle}{Proceedings of the AAAI conference on artificial intelligence}, pp. \bibinfo{pages}{7464--7471}.
\bibitem[{Cao et~al.(2018)Cao, Wei, Li and Li}]{cao2018faithful}
\bibinfo{author}{Cao, Z.}, \bibinfo{author}{Wei, F.}, \bibinfo{author}{Li, W.}, \bibinfo{author}{Li, S.}, \bibinfo{year}{2018}.
\newblock \bibinfo{title}{Faithful to the original: Fact aware neural abstractive summarization}, in: \bibinfo{booktitle}{Proceedings of the AAAI Conference on Artificial Intelligence}.
\bibitem[{Carlson et~al.(2010)Carlson, Betteridge, Kisiel, Settles, Hruschka and Mitchell}]{carlson2010toward}
\bibinfo{author}{Carlson, A.}, \bibinfo{author}{Betteridge, J.}, \bibinfo{author}{Kisiel, B.}, \bibinfo{author}{Settles, B.}, \bibinfo{author}{Hruschka, E.}, \bibinfo{author}{Mitchell, T.}, \bibinfo{year}{2010}.
\newblock \bibinfo{title}{Toward an architecture for never-ending language learning}, in: \bibinfo{booktitle}{Proceedings of the AAAI conference on artificial intelligence}, pp. \bibinfo{pages}{1306--1313}.
\bibitem[{Chen et~al.(2018)Chen, Wang, Utiyama, Sumita and Zhao}]{chen2018syntax}
\bibinfo{author}{Chen, K.}, \bibinfo{author}{Wang, R.}, \bibinfo{author}{Utiyama, M.}, \bibinfo{author}{Sumita, E.}, \bibinfo{author}{Zhao, T.}, \bibinfo{year}{2018}.
\newblock \bibinfo{title}{Syntax-directed attention for neural machine translation}, in: \bibinfo{booktitle}{Proceedings of the AAAI conference on artificial intelligence}.
\bibitem[{Chen et~al.(2019)Chen, Wu and Zaki}]{chen2019reinforcement}
\bibinfo{author}{Chen, Y.}, \bibinfo{author}{Wu, L.}, \bibinfo{author}{Zaki, M.J.}, \bibinfo{year}{2019}.
\newblock \bibinfo{title}{Reinforcement learning based graph-to-sequence model for natural question generation}.
\newblock \bibinfo{journal}{arXiv preprint arXiv:1908.04942} .
\bibitem[{Chen et~al.(2023)Chen, Wu and Zaki}]{chen2023toward}
\bibinfo{author}{Chen, Y.}, \bibinfo{author}{Wu, L.}, \bibinfo{author}{Zaki, M.J.}, \bibinfo{year}{2023}.
\newblock \bibinfo{title}{Toward subgraph-guided knowledge graph question generation with graph neural networks}.
\newblock \bibinfo{journal}{IEEE Transactions on Neural Networks and Learning Systems} .
\bibitem[{Ciampaglia et~al.(2015)Ciampaglia, Shiralkar, Rocha, Bollen, Menczer and Flammini}]{ciampaglia2015computational}
\bibinfo{author}{Ciampaglia, G.L.}, \bibinfo{author}{Shiralkar, P.}, \bibinfo{author}{Rocha, L.M.}, \bibinfo{author}{Bollen, J.}, \bibinfo{author}{Menczer, F.}, \bibinfo{author}{Flammini, A.}, \bibinfo{year}{2015}.
\newblock \bibinfo{title}{Computational fact checking from knowledge networks}.
\newblock \bibinfo{journal}{PloS one} \bibinfo{volume}{10}, \bibinfo{pages}{e0128193}.
\bibitem[{Dale(2000)}]{dale2000symbolic}
\bibinfo{author}{Dale, R.}, \bibinfo{year}{2000}.
\newblock \bibinfo{title}{Symbolic approaches to natural language processing}.
\newblock \bibinfo{journal}{Handbook of Natural Language Processing} , \bibinfo{pages}{1--9}.
\bibitem[{Devlin et~al.(2019)Devlin, Chang, Lee and Toutanova}]{bert}
\bibinfo{author}{Devlin, J.}, \bibinfo{author}{Chang, M.W.}, \bibinfo{author}{Lee, K.}, \bibinfo{author}{Toutanova, K.}, \bibinfo{year}{2019}.
\newblock \bibinfo{title}{{BERT}: Pre-training of deep bidirectional transformers for language understanding}, in: \bibinfo{editor}{Burstein, J.}, \bibinfo{editor}{Doran, C.}, \bibinfo{editor}{Solorio, T.} (Eds.), \bibinfo{booktitle}{Proceedings of the 2019 Conference of the North {A}merican Chapter of the Association for Computational Linguistics: Human Language Technologies, Volume 1 (Long and Short Papers)}, \bibinfo{publisher}{Association for Computational Linguistics}, \bibinfo{address}{Minneapolis, Minnesota}. pp. \bibinfo{pages}{4171--4186}.
\newblock \URLprefix \url{https://aclanthology.org/N19-1423}, \DOIprefix\doi{10.18653/v1/N19-1423}.
\bibitem[{Dinan et~al.(2018)Dinan, Roller, Shuster, Fan, Auli and Weston}]{dinan2018wizard}
\bibinfo{author}{Dinan, E.}, \bibinfo{author}{Roller, S.}, \bibinfo{author}{Shuster, K.}, \bibinfo{author}{Fan, A.}, \bibinfo{author}{Auli, M.}, \bibinfo{author}{Weston, J.}, \bibinfo{year}{2018}.
\newblock \bibinfo{title}{Wizard of wikipedia: Knowledge-powered conversational agents}.
\newblock \bibinfo{journal}{arXiv preprint arXiv:1811.01241} .
\bibitem[{Du and Cardie(2018)}]{du-cardie-2018-harvesting}
\bibinfo{author}{Du, X.}, \bibinfo{author}{Cardie, C.}, \bibinfo{year}{2018}.
\newblock \bibinfo{title}{Harvesting paragraph-level question-answer pairs from {W}ikipedia}, in: \bibinfo{booktitle}{Proceedings of the 56th Annual Meeting of the Association for Computational Linguistics (Volume 1: Long Papers)}, \bibinfo{publisher}{Association for Computational Linguistics}, \bibinfo{address}{Melbourne, Australia}. pp. \bibinfo{pages}{1907--1917}.
\newblock \URLprefix \url{https://aclanthology.org/P18-1177}, \DOIprefix\doi{10.18653/v1/P18-1177}.
\bibitem[{El-Kassas et~al.(2021)El-Kassas, Salama, Rafea and Mohamed}]{el2021automatic}
\bibinfo{author}{El-Kassas, W.S.}, \bibinfo{author}{Salama, C.R.}, \bibinfo{author}{Rafea, A.A.}, \bibinfo{author}{Mohamed, H.K.}, \bibinfo{year}{2021}.
\newblock \bibinfo{title}{Automatic text summarization: A comprehensive survey}.
\newblock \bibinfo{journal}{Expert systems with applications} \bibinfo{volume}{165}, \bibinfo{pages}{113679}.
\bibitem[{El~Mekki et~al.(2022)El~Mekki, El~Mahdaouy, Berrada and Khoumsi}]{el2022adasl}
\bibinfo{author}{El~Mekki, A.}, \bibinfo{author}{El~Mahdaouy, A.}, \bibinfo{author}{Berrada, I.}, \bibinfo{author}{Khoumsi, A.}, \bibinfo{year}{2022}.
\newblock \bibinfo{title}{Adasl: an unsupervised domain adaptation framework for arabic multi-dialectal sequence labeling}.
\newblock \bibinfo{journal}{Information Processing \& Management} \bibinfo{volume}{59}, \bibinfo{pages}{102964}.
\bibitem[{Elsahar et~al.(2018)Elsahar, Gravier and Laforest}]{elsahar-etal-2018-zero}
\bibinfo{author}{Elsahar, H.}, \bibinfo{author}{Gravier, C.}, \bibinfo{author}{Laforest, F.}, \bibinfo{year}{2018}.
\newblock \bibinfo{title}{Zero-shot question generation from knowledge graphs for unseen predicates and entity types}, in: \bibinfo{booktitle}{Proceedings of the 2018 Conference of the North {A}merican Chapter of the Association for Computational Linguistics: Human Language Technologies, Volume 1 (Long Papers)}, \bibinfo{publisher}{Association for Computational Linguistics}, \bibinfo{address}{New Orleans, Louisiana}. pp. \bibinfo{pages}{218--228}.
\newblock \URLprefix \url{https://aclanthology.org/N18-1020}, \DOIprefix\doi{10.18653/v1/N18-1020}.
\bibitem[{Etzioni et~al.(2008)Etzioni, Banko, Soderland and Weld}]{etzioni2008open}
\bibinfo{author}{Etzioni, O.}, \bibinfo{author}{Banko, M.}, \bibinfo{author}{Soderland, S.}, \bibinfo{author}{Weld, D.S.}, \bibinfo{year}{2008}.
\newblock \bibinfo{title}{Open information extraction from the web}.
\newblock \bibinfo{journal}{Communications of the ACM} \bibinfo{volume}{51}, \bibinfo{pages}{68--74}.
\bibitem[{Fellbaum(2010)}]{fellbaum2010wordnet}
\bibinfo{author}{Fellbaum, C.}, \bibinfo{year}{2010}.
\newblock \bibinfo{title}{Wordnet}, in: \bibinfo{booktitle}{Theory and applications of ontology: computer applications}. \bibinfo{publisher}{Springer}, pp. \bibinfo{pages}{231--243}.
\bibitem[{Feng et~al.(2020)Feng, Chen, Lin, Wang, Yan and Ren}]{feng-etal-2020-scalable}
\bibinfo{author}{Feng, Y.}, \bibinfo{author}{Chen, X.}, \bibinfo{author}{Lin, B.Y.}, \bibinfo{author}{Wang, P.}, \bibinfo{author}{Yan, J.}, \bibinfo{author}{Ren, X.}, \bibinfo{year}{2020}.
\newblock \bibinfo{title}{Scalable multi-hop relational reasoning for knowledge-aware question answering}, in: \bibinfo{booktitle}{Proceedings of the 2020 Conference on Empirical Methods in Natural Language Processing (EMNLP)}, \bibinfo{publisher}{Association for Computational Linguistics}, \bibinfo{address}{Online}. pp. \bibinfo{pages}{1295--1309}.
\newblock \URLprefix \url{https://aclanthology.org/2020.emnlp-main.99}, \DOIprefix\doi{10.18653/v1/2020.emnlp-main.99}.
\bibitem[{Ferrada et~al.(2017)Ferrada, Bustos and Hogan}]{ferrada2017imgpedia}
\bibinfo{author}{Ferrada, S.}, \bibinfo{author}{Bustos, B.}, \bibinfo{author}{Hogan, A.}, \bibinfo{year}{2017}.
\newblock \bibinfo{title}{Imgpedia: a linked dataset with content-based analysis of wikimedia images}, in: \bibinfo{booktitle}{The Semantic Web--ISWC 2017: 16th International Semantic Web Conference, Vienna, Austria, October 21-25, 2017, Proceedings, Part II 16}, \bibinfo{organization}{Springer}. pp. \bibinfo{pages}{84--93}.
\bibitem[{Ferrone and Zanzotto(2020)}]{ferrone2020symbolic}
\bibinfo{author}{Ferrone, L.}, \bibinfo{author}{Zanzotto, F.M.}, \bibinfo{year}{2020}.
\newblock \bibinfo{title}{Symbolic, distributed, and distributional representations for natural language processing in the era of deep learning: A survey}.
\newblock \bibinfo{journal}{Frontiers in Robotics and AI} \bibinfo{volume}{6}, \bibinfo{pages}{153}.
\bibitem[{Fromm et~al.(2019)Fromm, Faerman and Seidl}]{fromm2019tacam}
\bibinfo{author}{Fromm, M.}, \bibinfo{author}{Faerman, E.}, \bibinfo{author}{Seidl, T.}, \bibinfo{year}{2019}.
\newblock \bibinfo{title}{Tacam: topic and context aware argument mining}, in: \bibinfo{booktitle}{IEEE/WIC/ACM International Conference on Web Intelligence}, pp. \bibinfo{pages}{99--106}.
\bibitem[{Ganea and Hofmann(2017)}]{ganea-hofmann-2017-deep}
\bibinfo{author}{Ganea, O.E.}, \bibinfo{author}{Hofmann, T.}, \bibinfo{year}{2017}.
\newblock \bibinfo{title}{Deep joint entity disambiguation with local neural attention}, in: \bibinfo{booktitle}{Proceedings of the 2017 Conference on Empirical Methods in Natural Language Processing}, \bibinfo{publisher}{Association for Computational Linguistics}, \bibinfo{address}{Copenhagen, Denmark}. pp. \bibinfo{pages}{2619--2629}.
\newblock \URLprefix \url{https://aclanthology.org/D17-1277}, \DOIprefix\doi{10.18653/v1/D17-1277}.
\bibitem[{Gardner et~al.(2017)Gardner, Grus, Neumann, Tafjord, Dasigi, Liu, Peters, Schmitz and Zettlemoyer}]{Gardner2017AllenNLP}
\bibinfo{author}{Gardner, M.}, \bibinfo{author}{Grus, J.}, \bibinfo{author}{Neumann, M.}, \bibinfo{author}{Tafjord, O.}, \bibinfo{author}{Dasigi, P.}, \bibinfo{author}{Liu, N.F.}, \bibinfo{author}{Peters, M.}, \bibinfo{author}{Schmitz, M.}, \bibinfo{author}{Zettlemoyer, L.S.}, \bibinfo{year}{2017}.
\newblock \bibinfo{title}{Allennlp: A deep semantic natural language processing platform}.
\newblock \href{http://arxiv.org/abs/arXiv:1803.07640}{{\tt arXiv:arXiv:1803.07640}}.
\bibitem[{Ghazvininejad et~al.(2018)Ghazvininejad, Brockett, Chang, Dolan, Gao, Yih and Galley}]{ghazvininejad2018knowledge}
\bibinfo{author}{Ghazvininejad, M.}, \bibinfo{author}{Brockett, C.}, \bibinfo{author}{Chang, M.W.}, \bibinfo{author}{Dolan, B.}, \bibinfo{author}{Gao, J.}, \bibinfo{author}{Yih, W.t.}, \bibinfo{author}{Galley, M.}, \bibinfo{year}{2018}.
\newblock \bibinfo{title}{A knowledge-grounded neural conversation model}, in: \bibinfo{booktitle}{Proceedings of the AAAI Conference on Artificial Intelligence}.
\bibitem[{Grover and Leskovec(2016)}]{grover2016node2vec}
\bibinfo{author}{Grover, A.}, \bibinfo{author}{Leskovec, J.}, \bibinfo{year}{2016}.
\newblock \bibinfo{title}{node2vec: Scalable feature learning for networks}, in: \bibinfo{booktitle}{Proceedings of the 22nd ACM SIGKDD international conference on Knowledge discovery and data mining}, pp. \bibinfo{pages}{855--864}.
\bibitem[{Gu et~al.(2023)Gu, Wang, Zhang, Wu and Gu}]{gu2023enhancing}
\bibinfo{author}{Gu, Y.}, \bibinfo{author}{Wang, Y.}, \bibinfo{author}{Zhang, H.R.}, \bibinfo{author}{Wu, J.}, \bibinfo{author}{Gu, X.}, \bibinfo{year}{2023}.
\newblock \bibinfo{title}{Enhancing text classification by graph neural networks with multi-granular topic-aware graph}.
\newblock \bibinfo{journal}{IEEE Access} \bibinfo{volume}{11}, \bibinfo{pages}{20169--20183}.
\bibitem[{Guan et~al.(2020)Guan, Huang, Zhao, Zhu and Huang}]{guan2020knowledge}
\bibinfo{author}{Guan, J.}, \bibinfo{author}{Huang, F.}, \bibinfo{author}{Zhao, Z.}, \bibinfo{author}{Zhu, X.}, \bibinfo{author}{Huang, M.}, \bibinfo{year}{2020}.
\newblock \bibinfo{title}{A knowledge-enhanced pretraining model for commonsense story generation}.
\newblock \bibinfo{journal}{Transactions of the Association for Computational Linguistics} \bibinfo{volume}{8}, \bibinfo{pages}{93--108}.
\bibitem[{Gui et~al.(2019)Gui, Zou, Zhang, Peng, Fu, Wei and Huang}]{gui2019lexicon}
\bibinfo{author}{Gui, T.}, \bibinfo{author}{Zou, Y.}, \bibinfo{author}{Zhang, Q.}, \bibinfo{author}{Peng, M.}, \bibinfo{author}{Fu, J.}, \bibinfo{author}{Wei, Z.}, \bibinfo{author}{Huang, X.J.}, \bibinfo{year}{2019}.
\newblock \bibinfo{title}{A lexicon-based graph neural network for chinese ner}, in: \bibinfo{booktitle}{Proceedings of the 2019 conference on empirical methods in natural language processing and the 9th international joint conference on natural language processing (EMNLP-IJCNLP)}, pp. \bibinfo{pages}{1040--1050}.
\bibitem[{Guo et~al.(2022)Guo, Schlichtkrull and Vlachos}]{guo2022survey}
\bibinfo{author}{Guo, Z.}, \bibinfo{author}{Schlichtkrull, M.}, \bibinfo{author}{Vlachos, A.}, \bibinfo{year}{2022}.
\newblock \bibinfo{title}{A survey on automated fact-checking}.
\newblock \bibinfo{journal}{Transactions of the Association for Computational Linguistics} \bibinfo{volume}{10}, \bibinfo{pages}{178--206}.
\bibitem[{Hamilton et~al.(2022)Hamilton, Nayak, Bo{\v{z}}i{\'c} and Longo}]{hamilton2022neuro}
\bibinfo{author}{Hamilton, K.}, \bibinfo{author}{Nayak, A.}, \bibinfo{author}{Bo{\v{z}}i{\'c}, B.}, \bibinfo{author}{Longo, L.}, \bibinfo{year}{2022}.
\newblock \bibinfo{title}{Is neuro-symbolic ai meeting its promises in natural language processing? a structured review}.
\newblock \bibinfo{journal}{Semantic Web} , \bibinfo{pages}{1--42}.
\bibitem[{Han et~al.(2018)Han, Cao, Lv, Lin, Liu, Sun and Li}]{han-etal-2018-openke}
\bibinfo{author}{Han, X.}, \bibinfo{author}{Cao, S.}, \bibinfo{author}{Lv, X.}, \bibinfo{author}{Lin, Y.}, \bibinfo{author}{Liu, Z.}, \bibinfo{author}{Sun, M.}, \bibinfo{author}{Li, J.}, \bibinfo{year}{2018}.
\newblock \bibinfo{title}{{O}pen{KE}: An open toolkit for knowledge embedding}, in: \bibinfo{booktitle}{Proceedings of the 2018 Conference on Empirical Methods in Natural Language Processing: System Demonstrations}, \bibinfo{publisher}{Association for Computational Linguistics}, \bibinfo{address}{Brussels, Belgium}. pp. \bibinfo{pages}{139--144}.
\newblock \URLprefix \url{https://aclanthology.org/D18-2024}, \DOIprefix\doi{10.18653/v1/D18-2024}.
\bibitem[{Hoehndorf et~al.(2017)Hoehndorf, Queralt-Rosinach et~al.}]{hoehndorf2017data}
\bibinfo{author}{Hoehndorf, R.}, \bibinfo{author}{Queralt-Rosinach, N.}, et~al., \bibinfo{year}{2017}.
\newblock \bibinfo{title}{Data science and symbolic ai: Synergies, challenges and opportunities}.
\newblock \bibinfo{journal}{Data Science} \bibinfo{volume}{1}, \bibinfo{pages}{27--38}.
\bibitem[{Hou et~al.(2020)Hou, Zhang and Fei}]{hou2020rhetorical}
\bibinfo{author}{Hou, S.}, \bibinfo{author}{Zhang, S.}, \bibinfo{author}{Fei, C.}, \bibinfo{year}{2020}.
\newblock \bibinfo{title}{Rhetorical structure theory: A comprehensive review of theory, parsing methods and applications}.
\newblock \bibinfo{journal}{Expert Systems with Applications} \bibinfo{volume}{157}, \bibinfo{pages}{113421}.
\bibitem[{Hu et~al.(2023)Hu, Liu, Zhao, Hou, Nie and Li}]{hu2023survey}
\bibinfo{author}{Hu, L.}, \bibinfo{author}{Liu, Z.}, \bibinfo{author}{Zhao, Z.}, \bibinfo{author}{Hou, L.}, \bibinfo{author}{Nie, L.}, \bibinfo{author}{Li, J.}, \bibinfo{year}{2023}.
\newblock \bibinfo{title}{A survey of knowledge enhanced pre-trained language models}.
\newblock \bibinfo{journal}{IEEE Transactions on Knowledge and Data Engineering} .
\bibitem[{Hu et~al.(2022)Hu, He, Ma, Wang and Xiao}]{hu2022kgner}
\bibinfo{author}{Hu, W.}, \bibinfo{author}{He, L.}, \bibinfo{author}{Ma, H.}, \bibinfo{author}{Wang, K.}, \bibinfo{author}{Xiao, J.}, \bibinfo{year}{2022}.
\newblock \bibinfo{title}{Kgner: Improving chinese named entity recognition by bert infused with the knowledge graph}.
\newblock \bibinfo{journal}{Applied Sciences} \bibinfo{volume}{12}, \bibinfo{pages}{7702}.
\bibitem[{Huo et~al.(2022)Huo, Ma and Liu}]{HUO2022102980}
\bibinfo{author}{Huo, C.}, \bibinfo{author}{Ma, S.}, \bibinfo{author}{Liu, X.}, \bibinfo{year}{2022}.
\newblock \bibinfo{title}{Hotness prediction of scientific topics based on a bibliographic knowledge graph}.
\newblock \bibinfo{journal}{Information Processing \& Management} \bibinfo{volume}{59}, \bibinfo{pages}{102980}.
\newblock \URLprefix \url{https://www.sciencedirect.com/science/article/pii/S0306457322000966}, \DOIprefix\doi{https://doi.org/10.1016/j.ipm.2022.102980}.
\bibitem[{Hwang et~al.(2021)Hwang, Bhagavatula, Le~Bras, Da, Sakaguchi, Bosselut and Choi}]{hwang2021comet}
\bibinfo{author}{Hwang, J.D.}, \bibinfo{author}{Bhagavatula, C.}, \bibinfo{author}{Le~Bras, R.}, \bibinfo{author}{Da, J.}, \bibinfo{author}{Sakaguchi, K.}, \bibinfo{author}{Bosselut, A.}, \bibinfo{author}{Choi, Y.}, \bibinfo{year}{2021}.
\newblock \bibinfo{title}{(comet-) atomic 2020: on symbolic and neural commonsense knowledge graphs}, in: \bibinfo{booktitle}{Proceedings of the AAAI Conference on Artificial Intelligence}, pp. \bibinfo{pages}{6384--6392}.
\bibitem[{Ji et~al.(2020)Ji, Ke, Huang, Wei, Zhu and Huang}]{ji-etal-2020-language}
\bibinfo{author}{Ji, H.}, \bibinfo{author}{Ke, P.}, \bibinfo{author}{Huang, S.}, \bibinfo{author}{Wei, F.}, \bibinfo{author}{Zhu, X.}, \bibinfo{author}{Huang, M.}, \bibinfo{year}{2020}.
\newblock \bibinfo{title}{Language generation with multi-hop reasoning on commonsense knowledge graph}, in: \bibinfo{booktitle}{Proceedings of the 2020 Conference on Empirical Methods in Natural Language Processing (EMNLP)}, \bibinfo{publisher}{Association for Computational Linguistics}, \bibinfo{address}{Online}. pp. \bibinfo{pages}{725--736}.
\newblock \URLprefix \url{https://aclanthology.org/2020.emnlp-main.54}, \DOIprefix\doi{10.18653/v1/2020.emnlp-main.54}.
\bibitem[{Jia et~al.(2016)Jia, Wang, Lin, Jin and Cheng}]{jia2016locally}
\bibinfo{author}{Jia, Y.}, \bibinfo{author}{Wang, Y.}, \bibinfo{author}{Lin, H.}, \bibinfo{author}{Jin, X.}, \bibinfo{author}{Cheng, X.}, \bibinfo{year}{2016}.
\newblock \bibinfo{title}{Locally adaptive translation for knowledge graph embedding}, in: \bibinfo{booktitle}{Proceedings of the AAAI Conference on Artificial Intelligence}.
\bibitem[{Kalyan et~al.(2021)Kalyan, Rajasekharan and Sangeetha}]{xlm-r}
\bibinfo{author}{Kalyan, K.S.}, \bibinfo{author}{Rajasekharan, A.}, \bibinfo{author}{Sangeetha, S.}, \bibinfo{year}{2021}.
\newblock \bibinfo{title}{Ammus: A survey of transformer-based pretrained models in natural language processing}.
\newblock \bibinfo{journal}{arXiv preprint arXiv:2108.05542} .
\bibitem[{Karpukhin et~al.(2020)Karpukhin, Oguz, Min, Lewis, Wu, Edunov, Chen and Yih}]{karpukhin-etal-2020-dense}
\bibinfo{author}{Karpukhin, V.}, \bibinfo{author}{Oguz, B.}, \bibinfo{author}{Min, S.}, \bibinfo{author}{Lewis, P.}, \bibinfo{author}{Wu, L.}, \bibinfo{author}{Edunov, S.}, \bibinfo{author}{Chen, D.}, \bibinfo{author}{Yih, W.t.}, \bibinfo{year}{2020}.
\newblock \bibinfo{title}{Dense passage retrieval for open-domain question answering}, in: \bibinfo{booktitle}{Proceedings of the 2020 Conference on Empirical Methods in Natural Language Processing (EMNLP)}, \bibinfo{publisher}{Association for Computational Linguistics}, \bibinfo{address}{Online}. pp. \bibinfo{pages}{6769--6781}.
\newblock \URLprefix \url{https://aclanthology.org/2020.emnlp-main.550}, \DOIprefix\doi{10.18653/v1/2020.emnlp-main.550}.
\bibitem[{Kazemi et~al.(2021)Kazemi, Garimella, Gaffney and Hale}]{kazemi2021claim}
\bibinfo{author}{Kazemi, A.}, \bibinfo{author}{Garimella, K.}, \bibinfo{author}{Gaffney, D.}, \bibinfo{author}{Hale, S.}, \bibinfo{year}{2021}.
\newblock \bibinfo{title}{Claim matching beyond english to scale global fact-checking}, in: \bibinfo{booktitle}{Proceedings of the 59th Annual Meeting of the Association for Computational Linguistics and the 11th International Joint Conference on Natural Language Processing (Volume 1: Long Papers)}, pp. \bibinfo{pages}{4504--4517}.
\bibitem[{Koncel-Kedziorski et~al.(2019)Koncel-Kedziorski, Bekal, Luan, Lapata and Hajishirzi}]{koncel-kedziorski-etal-2019-text}
\bibinfo{author}{Koncel-Kedziorski, R.}, \bibinfo{author}{Bekal, D.}, \bibinfo{author}{Luan, Y.}, \bibinfo{author}{Lapata, M.}, \bibinfo{author}{Hajishirzi, H.}, \bibinfo{year}{2019}.
\newblock \bibinfo{title}{{T}ext {G}eneration from {K}nowledge {G}raphs with {G}raph {T}ransformers}, in: \bibinfo{booktitle}{Proceedings of the 2019 Conference of the North {A}merican Chapter of the Association for Computational Linguistics: Human Language Technologies, Volume 1 (Long and Short Papers)}, \bibinfo{publisher}{Association for Computational Linguistics}, \bibinfo{address}{Minneapolis, Minnesota}. pp. \bibinfo{pages}{2284--2293}.
\newblock \URLprefix \url{https://aclanthology.org/N19-1238}, \DOIprefix\doi{10.18653/v1/N19-1238}.
\bibitem[{Kumar et~al.(2019)Kumar, Hua, Ramakrishnan, Qi, Gao and Li}]{kumar2019difficulty}
\bibinfo{author}{Kumar, V.}, \bibinfo{author}{Hua, Y.}, \bibinfo{author}{Ramakrishnan, G.}, \bibinfo{author}{Qi, G.}, \bibinfo{author}{Gao, L.}, \bibinfo{author}{Li, Y.F.}, \bibinfo{year}{2019}.
\newblock \bibinfo{title}{Difficulty-controllable multi-hop question generation from knowledge graphs}, in: \bibinfo{booktitle}{The Semantic Web--ISWC 2019: 18th International Semantic Web Conference, Auckland, New Zealand, October 26--30, 2019, Proceedings, Part I 18}, \bibinfo{organization}{Springer}. pp. \bibinfo{pages}{382--398}.
\bibitem[{Kurdi et~al.(2020)Kurdi, Leo, Parsia, Sattler and Al-Emari}]{kurdi2020systematic}
\bibinfo{author}{Kurdi, G.}, \bibinfo{author}{Leo, J.}, \bibinfo{author}{Parsia, B.}, \bibinfo{author}{Sattler, U.}, \bibinfo{author}{Al-Emari, S.}, \bibinfo{year}{2020}.
\newblock \bibinfo{title}{A systematic review of automatic question generation for educational purposes}.
\newblock \bibinfo{journal}{International Journal of Artificial Intelligence in Education} \bibinfo{volume}{30}, \bibinfo{pages}{121--204}.
\bibitem[{Lawrence and Reed(2019)}]{lawrence-reed-2019-argument}
\bibinfo{author}{Lawrence, J.}, \bibinfo{author}{Reed, C.}, \bibinfo{year}{2019}.
\newblock \bibinfo{title}{Argument mining: A survey}.
\newblock \bibinfo{journal}{Computational Linguistics} \bibinfo{volume}{45}, \bibinfo{pages}{765--818}.
\newblock \URLprefix \url{https://aclanthology.org/J19-4006}, \DOIprefix\doi{10.1162/coli_a_00364}.
\bibitem[{Levy et~al.(2017)Levy, Seo, Choi and Zettlemoyer}]{levy2017zero}
\bibinfo{author}{Levy, O.}, \bibinfo{author}{Seo, M.}, \bibinfo{author}{Choi, E.}, \bibinfo{author}{Zettlemoyer, L.}, \bibinfo{year}{2017}.
\newblock \bibinfo{title}{Zero-shot relation extraction via reading comprehension}, in: \bibinfo{booktitle}{Proceedings of the 21st Conference on Computational Natural Language Learning (CoNLL 2017)}, pp. \bibinfo{pages}{333--342}.
\bibitem[{Lewis et~al.(2020)Lewis, Liu, Goyal, Ghazvininejad, Mohamed, Levy, Stoyanov and Zettlemoyer}]{bart}
\bibinfo{author}{Lewis, M.}, \bibinfo{author}{Liu, Y.}, \bibinfo{author}{Goyal, N.}, \bibinfo{author}{Ghazvininejad, M.}, \bibinfo{author}{Mohamed, A.}, \bibinfo{author}{Levy, O.}, \bibinfo{author}{Stoyanov, V.}, \bibinfo{author}{Zettlemoyer, L.}, \bibinfo{year}{2020}.
\newblock \bibinfo{title}{{BART}: Denoising sequence-to-sequence pre-training for natural language generation, translation, and comprehension}, in: \bibinfo{editor}{Jurafsky, D.}, \bibinfo{editor}{Chai, J.}, \bibinfo{editor}{Schluter, N.}, \bibinfo{editor}{Tetreault, J.} (Eds.), \bibinfo{booktitle}{Proceedings of the 58th Annual Meeting of the Association for Computational Linguistics}, \bibinfo{publisher}{Association for Computational Linguistics}, \bibinfo{address}{Online}. pp. \bibinfo{pages}{7871--7880}.
\newblock \URLprefix \url{https://aclanthology.org/2020.acl-main.703}, \DOIprefix\doi{10.18653/v1/2020.acl-main.703}.
\bibitem[{Li et~al.(2018)Li, Zhu, Zhang and Zong}]{li-etal-2018-ensure}
\bibinfo{author}{Li, H.}, \bibinfo{author}{Zhu, J.}, \bibinfo{author}{Zhang, J.}, \bibinfo{author}{Zong, C.}, \bibinfo{year}{2018}.
\newblock \bibinfo{title}{Ensure the correctness of the summary: Incorporate entailment knowledge into abstractive sentence summarization}, in: \bibinfo{booktitle}{Proceedings of the 27th International Conference on Computational Linguistics}, \bibinfo{publisher}{Association for Computational Linguistics}, \bibinfo{address}{Santa Fe, New Mexico, USA}. pp. \bibinfo{pages}{1430--1441}.
\newblock \URLprefix \url{https://aclanthology.org/C18-1121}.
\bibitem[{Li et~al.(2022)Li, Chen, Dong, Keutzer and Zhang}]{lidomain}
\bibinfo{author}{Li, T.}, \bibinfo{author}{Chen, X.}, \bibinfo{author}{Dong, Z.}, \bibinfo{author}{Keutzer, K.}, \bibinfo{author}{Zhang, S.}, \bibinfo{year}{2022}.
\newblock \bibinfo{title}{Domain-adaptive text classification with structured knowledge from unlabeled data}, \bibinfo{organization}{IJCAI International Joint Conference on Artificial Intelligence}.
\bibitem[{Li et~al.(2021)Li, Ding, Liu, Hu and Van~Durme}]{li2021guided}
\bibinfo{author}{Li, Z.}, \bibinfo{author}{Ding, X.}, \bibinfo{author}{Liu, T.}, \bibinfo{author}{Hu, J.E.}, \bibinfo{author}{Van~Durme, B.}, \bibinfo{year}{2021}.
\newblock \bibinfo{title}{Guided generation of cause and effect}.
\newblock \bibinfo{journal}{arXiv preprint arXiv:2107.09846} .
\bibitem[{Liu et~al.(2020a)Liu, Zhou, Zhao, Wang, Ju, Deng and Wang}]{liu2020k}
\bibinfo{author}{Liu, W.}, \bibinfo{author}{Zhou, P.}, \bibinfo{author}{Zhao, Z.}, \bibinfo{author}{Wang, Z.}, \bibinfo{author}{Ju, Q.}, \bibinfo{author}{Deng, H.}, \bibinfo{author}{Wang, P.}, \bibinfo{year}{2020}a.
\newblock \bibinfo{title}{K-bert: Enabling language representation with knowledge graph}, in: \bibinfo{booktitle}{Proceedings of the AAAI Conference on Artificial Intelligence}, pp. \bibinfo{pages}{2901--2908}.
\bibitem[{Liu et~al.(2019a)Liu, Li, Garcia-Duran, Niepert, Onoro-Rubio and Rosenblum}]{liu2019mmkg}
\bibinfo{author}{Liu, Y.}, \bibinfo{author}{Li, H.}, \bibinfo{author}{Garcia-Duran, A.}, \bibinfo{author}{Niepert, M.}, \bibinfo{author}{Onoro-Rubio, D.}, \bibinfo{author}{Rosenblum, D.S.}, \bibinfo{year}{2019}a.
\newblock \bibinfo{title}{Mmkg: multi-modal knowledge graphs}, in: \bibinfo{booktitle}{The Semantic Web: 16th International Conference, ESWC 2019, Portoro{\v{z}}, Slovenia, June 2--6, 2019, Proceedings 16}, \bibinfo{organization}{Springer}. pp. \bibinfo{pages}{459--474}.
\bibitem[{Liu et~al.(2022)Liu, Li and Hu}]{liu2022combining}
\bibinfo{author}{Liu, Y.}, \bibinfo{author}{Li, P.}, \bibinfo{author}{Hu, X.}, \bibinfo{year}{2022}.
\newblock \bibinfo{title}{Combining context-relevant features with multi-stage attention network for short text classification}.
\newblock \bibinfo{journal}{Computer Speech \& Language} \bibinfo{volume}{71}, \bibinfo{pages}{101268}.
\bibitem[{Liu et~al.(2019b)Liu, Ott, Goyal, Du, Joshi, Chen, Levy, Lewis, Zettlemoyer and Stoyanov}]{liu2019roberta}
\bibinfo{author}{Liu, Y.}, \bibinfo{author}{Ott, M.}, \bibinfo{author}{Goyal, N.}, \bibinfo{author}{Du, J.}, \bibinfo{author}{Joshi, M.}, \bibinfo{author}{Chen, D.}, \bibinfo{author}{Levy, O.}, \bibinfo{author}{Lewis, M.}, \bibinfo{author}{Zettlemoyer, L.}, \bibinfo{author}{Stoyanov, V.}, \bibinfo{year}{2019}b.
\newblock \bibinfo{title}{Roberta: A robustly optimized bert pretraining approach}.
\newblock \bibinfo{journal}{arXiv preprint arXiv:1907.11692} .
\bibitem[{Liu et~al.(2021)Liu, Wan, He, Peng and Philip}]{liu2021kg}
\bibinfo{author}{Liu, Y.}, \bibinfo{author}{Wan, Y.}, \bibinfo{author}{He, L.}, \bibinfo{author}{Peng, H.}, \bibinfo{author}{Philip, S.Y.}, \bibinfo{year}{2021}.
\newblock \bibinfo{title}{Kg-bart: Knowledge graph-augmented bart for generative commonsense reasoning}, in: \bibinfo{booktitle}{Proceedings of the AAAI Conference on Artificial Intelligence}, pp. \bibinfo{pages}{6418--6425}.
\bibitem[{Liu et~al.(2019c)Liu, Zeng, Ordieres~Mer{\'e} and Yang}]{liu2019anticipating}
\bibinfo{author}{Liu, Y.}, \bibinfo{author}{Zeng, Q.}, \bibinfo{author}{Ordieres~Mer{\'e}, J.}, \bibinfo{author}{Yang, H.}, \bibinfo{year}{2019}c.
\newblock \bibinfo{title}{Anticipating stock market of the renowned companies: A knowledge graph approach}.
\newblock \bibinfo{journal}{Complexity} \bibinfo{volume}{2019}.
\bibitem[{Liu et~al.(2019d)Liu, Niu, Wu and Wang}]{liu2019knowledge}
\bibinfo{author}{Liu, Z.}, \bibinfo{author}{Niu, Z.Y.}, \bibinfo{author}{Wu, H.}, \bibinfo{author}{Wang, H.}, \bibinfo{year}{2019}d.
\newblock \bibinfo{title}{Knowledge aware conversation generation with explainable reasoning over augmented graphs}, in: \bibinfo{booktitle}{Proceedings of the 2019 Conference on Empirical Methods in Natural Language Processing and the 9th International Joint Conference on Natural Language Processing (EMNLP-IJCNLP)}, pp. \bibinfo{pages}{1782--1792}.
\bibitem[{Liu et~al.(2020b)Liu, Xiong, Sun and Liu}]{liu-etal-2020-fine}
\bibinfo{author}{Liu, Z.}, \bibinfo{author}{Xiong, C.}, \bibinfo{author}{Sun, M.}, \bibinfo{author}{Liu, Z.}, \bibinfo{year}{2020}b.
\newblock \bibinfo{title}{Fine-grained fact verification with kernel graph attention network}, in: \bibinfo{booktitle}{Proceedings of the 58th Annual Meeting of the Association for Computational Linguistics}, \bibinfo{publisher}{Association for Computational Linguistics}, \bibinfo{address}{Online}. pp. \bibinfo{pages}{7342--7351}.
\newblock \URLprefix \url{https://aclanthology.org/2020.acl-main.655}, \DOIprefix\doi{10.18653/v1/2020.acl-main.655}.
\bibitem[{Lopez(2008)}]{lopez2008statistical}
\bibinfo{author}{Lopez, A.}, \bibinfo{year}{2008}.
\newblock \bibinfo{title}{Statistical machine translation}.
\newblock \bibinfo{journal}{ACM Computing Surveys (CSUR)} \bibinfo{volume}{40}, \bibinfo{pages}{1--49}.
\bibitem[{Lv et~al.(2020)Lv, Guo, Xu, Tang, Duan, Gong, Shou, Jiang, Cao and Hu}]{lv2020graph}
\bibinfo{author}{Lv, S.}, \bibinfo{author}{Guo, D.}, \bibinfo{author}{Xu, J.}, \bibinfo{author}{Tang, D.}, \bibinfo{author}{Duan, N.}, \bibinfo{author}{Gong, M.}, \bibinfo{author}{Shou, L.}, \bibinfo{author}{Jiang, D.}, \bibinfo{author}{Cao, G.}, \bibinfo{author}{Hu, S.}, \bibinfo{year}{2020}.
\newblock \bibinfo{title}{Graph-based reasoning over heterogeneous external knowledge for commonsense question answering}, in: \bibinfo{booktitle}{Proceedings of the AAAI conference on artificial intelligence}, pp. \bibinfo{pages}{8449--8456}.
\bibitem[{Meng et~al.(2020)Meng, Ren, Chen, Monz, Ma and de~Rijke}]{meng2020refnet}
\bibinfo{author}{Meng, C.}, \bibinfo{author}{Ren, P.}, \bibinfo{author}{Chen, Z.}, \bibinfo{author}{Monz, C.}, \bibinfo{author}{Ma, J.}, \bibinfo{author}{de~Rijke, M.}, \bibinfo{year}{2020}.
\newblock \bibinfo{title}{Refnet: A reference-aware network for background based conversation}, in: \bibinfo{booktitle}{Proceedings of the AAAI conference on artificial intelligence}, pp. \bibinfo{pages}{8496--8503}.
\bibitem[{Mikolov et~al.(2013)Mikolov, Chen, Corrado and Dean}]{mikolov2013efficient}
\bibinfo{author}{Mikolov, T.}, \bibinfo{author}{Chen, K.}, \bibinfo{author}{Corrado, G.}, \bibinfo{author}{Dean, J.}, \bibinfo{year}{2013}.
\newblock \bibinfo{title}{Efficient estimation of word representations in vector space}.
\newblock \bibinfo{journal}{arXiv preprint arXiv:1301.3781} .
\bibitem[{Min et~al.(2023)Min, Ross, Sulem, Veyseh, Nguyen, Sainz, Agirre, Heintz and Roth}]{min2023recent}
\bibinfo{author}{Min, B.}, \bibinfo{author}{Ross, H.}, \bibinfo{author}{Sulem, E.}, \bibinfo{author}{Veyseh, A.P.B.}, \bibinfo{author}{Nguyen, T.H.}, \bibinfo{author}{Sainz, O.}, \bibinfo{author}{Agirre, E.}, \bibinfo{author}{Heintz, I.}, \bibinfo{author}{Roth, D.}, \bibinfo{year}{2023}.
\newblock \bibinfo{title}{Recent advances in natural language processing via large pre-trained language models: A survey}.
\newblock \bibinfo{journal}{ACM Computing Surveys} \bibinfo{volume}{56}, \bibinfo{pages}{1--40}.
\bibitem[{Mitra et~al.(2019)Mitra, Banerjee, Pal, Mishra and Baral}]{mitra2019additional}
\bibinfo{author}{Mitra, A.}, \bibinfo{author}{Banerjee, P.}, \bibinfo{author}{Pal, K.K.}, \bibinfo{author}{Mishra, S.}, \bibinfo{author}{Baral, C.}, \bibinfo{year}{2019}.
\newblock \bibinfo{title}{How additional knowledge can improve natural language commonsense question answering?}
\newblock \bibinfo{journal}{arXiv preprint arXiv:1909.08855} .
\bibitem[{Mulla and Gharpure(2023)}]{mulla2023automatic}
\bibinfo{author}{Mulla, N.}, \bibinfo{author}{Gharpure, P.}, \bibinfo{year}{2023}.
\newblock \bibinfo{title}{Automatic question generation: a review of methodologies, datasets, evaluation metrics, and applications}.
\newblock \bibinfo{journal}{Progress in Artificial Intelligence} \bibinfo{volume}{12}, \bibinfo{pages}{1--32}.
\bibitem[{Narayan et~al.(2018)Narayan, Cohen and Lapata}]{narayan-etal-2018-dont}
\bibinfo{author}{Narayan, S.}, \bibinfo{author}{Cohen, S.B.}, \bibinfo{author}{Lapata, M.}, \bibinfo{year}{2018}.
\newblock \bibinfo{title}{Don{'}t give me the details, just the summary! topic-aware convolutional neural networks for extreme summarization}, in: \bibinfo{booktitle}{Proceedings of the 2018 Conference on Empirical Methods in Natural Language Processing}, \bibinfo{publisher}{Association for Computational Linguistics}, \bibinfo{address}{Brussels, Belgium}. pp. \bibinfo{pages}{1797--1807}.
\newblock \URLprefix \url{https://aclanthology.org/D18-1206}, \DOIprefix\doi{10.18653/v1/D18-1206}.
\bibitem[{Ni et~al.(2023)Ni, Young, Pandelea, Xue and Cambria}]{ni2023recent}
\bibinfo{author}{Ni, J.}, \bibinfo{author}{Young, T.}, \bibinfo{author}{Pandelea, V.}, \bibinfo{author}{Xue, F.}, \bibinfo{author}{Cambria, E.}, \bibinfo{year}{2023}.
\newblock \bibinfo{title}{Recent advances in deep learning based dialogue systems: A systematic survey}.
\newblock \bibinfo{journal}{Artificial intelligence review} \bibinfo{volume}{56}, \bibinfo{pages}{3055--3155}.
\bibitem[{Noraset et~al.(2021)Noraset, Lowphansirikul and Tuarob}]{NORASET2021102431}
\bibinfo{author}{Noraset, T.}, \bibinfo{author}{Lowphansirikul, L.}, \bibinfo{author}{Tuarob, S.}, \bibinfo{year}{2021}.
\newblock \bibinfo{title}{Wabiqa: A wikipedia-based thai question-answering system}.
\newblock \bibinfo{journal}{Information Processing \& Management} \bibinfo{volume}{58}, \bibinfo{pages}{102431}.
\newblock \URLprefix \url{https://www.sciencedirect.com/science/article/pii/S0306457320309249}, \DOIprefix\doi{https://doi.org/10.1016/j.ipm.2020.102431}.
\bibitem[{Pan et~al.(2023)Pan, Luo, Wang, Chen, Wang and Wu}]{pan2023unifying}
\bibinfo{author}{Pan, S.}, \bibinfo{author}{Luo, L.}, \bibinfo{author}{Wang, Y.}, \bibinfo{author}{Chen, C.}, \bibinfo{author}{Wang, J.}, \bibinfo{author}{Wu, X.}, \bibinfo{year}{2023}.
\newblock \bibinfo{title}{Unifying large language models and knowledge graphs: A roadmap}.
\newblock \bibinfo{journal}{arXiv preprint arXiv:2306.08302} .
\bibitem[{P{\'e}rez-Ag{\"u}era et~al.(2010)P{\'e}rez-Ag{\"u}era, Arroyo, Greenberg, Iglesias and Fresno}]{perez2010using}
\bibinfo{author}{P{\'e}rez-Ag{\"u}era, J.R.}, \bibinfo{author}{Arroyo, J.}, \bibinfo{author}{Greenberg, J.}, \bibinfo{author}{Iglesias, J.P.}, \bibinfo{author}{Fresno, V.}, \bibinfo{year}{2010}.
\newblock \bibinfo{title}{Using bm25f for semantic search}, in: \bibinfo{booktitle}{Proceedings of the 3rd international semantic search workshop}, pp. \bibinfo{pages}{1--8}.
\bibitem[{Peters et~al.(2019)Peters, Neumann, Logan, Schwartz, Joshi, Singh and Smith}]{peters-etal-2019-knowledge}
\bibinfo{author}{Peters, M.E.}, \bibinfo{author}{Neumann, M.}, \bibinfo{author}{Logan, R.}, \bibinfo{author}{Schwartz, R.}, \bibinfo{author}{Joshi, V.}, \bibinfo{author}{Singh, S.}, \bibinfo{author}{Smith, N.A.}, \bibinfo{year}{2019}.
\newblock \bibinfo{title}{Knowledge enhanced contextual word representations}, in: \bibinfo{booktitle}{Proceedings of the 2019 Conference on Empirical Methods in Natural Language Processing and the 9th International Joint Conference on Natural Language Processing (EMNLP-IJCNLP)}, \bibinfo{publisher}{Association for Computational Linguistics}, \bibinfo{address}{Hong Kong, China}. pp. \bibinfo{pages}{43--54}.
\newblock \URLprefix \url{https://aclanthology.org/D19-1005}, \DOIprefix\doi{10.18653/v1/D19-1005}.
\bibitem[{Petroni et~al.(2019)Petroni, Rockt{\"a}schel, Riedel, Lewis, Bakhtin, Wu and Miller}]{petroni-etal-2019-language}
\bibinfo{author}{Petroni, F.}, \bibinfo{author}{Rockt{\"a}schel, T.}, \bibinfo{author}{Riedel, S.}, \bibinfo{author}{Lewis, P.}, \bibinfo{author}{Bakhtin, A.}, \bibinfo{author}{Wu, Y.}, \bibinfo{author}{Miller, A.}, \bibinfo{year}{2019}.
\newblock \bibinfo{title}{Language models as knowledge bases?}, in: \bibinfo{booktitle}{Proceedings of the 2019 Conference on Empirical Methods in Natural Language Processing and the 9th International Joint Conference on Natural Language Processing (EMNLP-IJCNLP)}, \bibinfo{publisher}{Association for Computational Linguistics}, \bibinfo{address}{Hong Kong, China}. pp. \bibinfo{pages}{2463--2473}.
\newblock \URLprefix \url{https://aclanthology.org/D19-1250}, \DOIprefix\doi{10.18653/v1/D19-1250}.
\bibitem[{Pittaras et~al.(2023)Pittaras, Giannakopoulos, Stamatopoulos and Karkaletsis}]{pittaras2023content}
\bibinfo{author}{Pittaras, N.}, \bibinfo{author}{Giannakopoulos, G.}, \bibinfo{author}{Stamatopoulos, P.}, \bibinfo{author}{Karkaletsis, V.}, \bibinfo{year}{2023}.
\newblock \bibinfo{title}{Content-based and knowledge-enriched representations for classification across modalities: a survey}.
\newblock \bibinfo{journal}{ACM Computing Surveys} .
\bibitem[{Radford and Narasimhan(2018)}]{GPT}
\bibinfo{author}{Radford, A.}, \bibinfo{author}{Narasimhan, K.}, \bibinfo{year}{2018}.
\newblock \bibinfo{title}{Improving language understanding by generative pre-training}.
\newblock \URLprefix \url{https://api.semanticscholar.org/CorpusID:49313245}.
\bibitem[{Reddy et~al.(2017)Reddy, Raghu, Khapra and Joshi}]{reddy-etal-2017-generating}
\bibinfo{author}{Reddy, S.}, \bibinfo{author}{Raghu, D.}, \bibinfo{author}{Khapra, M.M.}, \bibinfo{author}{Joshi, S.}, \bibinfo{year}{2017}.
\newblock \bibinfo{title}{Generating natural language question-answer pairs from a knowledge graph using a {RNN} based question generation model}, in: \bibinfo{booktitle}{Proceedings of the 15th Conference of the {E}uropean Chapter of the Association for Computational Linguistics: Volume 1, Long Papers}, \bibinfo{publisher}{Association for Computational Linguistics}, \bibinfo{address}{Valencia, Spain}. pp. \bibinfo{pages}{376--385}.
\newblock \URLprefix \url{https://aclanthology.org/E17-1036}.
\bibitem[{Reimers and Gurevych(2019)}]{sentence_bert}
\bibinfo{author}{Reimers, N.}, \bibinfo{author}{Gurevych, I.}, \bibinfo{year}{2019}.
\newblock \bibinfo{title}{Sentence-bert: Sentence embeddings using siamese bert-networks}.
\newblock \bibinfo{journal}{arXiv preprint arXiv:1908.10084} .
\bibitem[{Ribeiro et~al.(2017)Ribeiro, Saverese and Figueiredo}]{ribeiro2017struc2vec}
\bibinfo{author}{Ribeiro, L.F.}, \bibinfo{author}{Saverese, P.H.}, \bibinfo{author}{Figueiredo, D.R.}, \bibinfo{year}{2017}.
\newblock \bibinfo{title}{struc2vec: Learning node representations from structural identity}, in: \bibinfo{booktitle}{Proceedings of the 23rd ACM SIGKDD international conference on knowledge discovery and data mining}, pp. \bibinfo{pages}{385--394}.
\bibitem[{Rosenfeld(2000)}]{rosenfeld2000two}
\bibinfo{author}{Rosenfeld, R.}, \bibinfo{year}{2000}.
\newblock \bibinfo{title}{Two decades of statistical language modeling: Where do we go from here?}
\newblock \bibinfo{journal}{Proceedings of the IEEE} \bibinfo{volume}{88}, \bibinfo{pages}{1270--1278}.
\bibitem[{Saadat-Yazdi et~al.(2022)Saadat-Yazdi, Li, Chausson, Belle, Ross, Pan and K{\"o}kciyan}]{saadat-yazdi-etal-2022-kevin}
\bibinfo{author}{Saadat-Yazdi, A.}, \bibinfo{author}{Li, X.}, \bibinfo{author}{Chausson, S.}, \bibinfo{author}{Belle, V.}, \bibinfo{author}{Ross, B.}, \bibinfo{author}{Pan, J.Z.}, \bibinfo{author}{K{\"o}kciyan, N.}, \bibinfo{year}{2022}.
\newblock \bibinfo{title}{{KEV}i{N}: A knowledge enhanced validity and novelty classifier for arguments}, in: \bibinfo{booktitle}{Proceedings of the 9th Workshop on Argument Mining}, \bibinfo{publisher}{International Conference on Computational Linguistics}, \bibinfo{address}{Online and in Gyeongju, Republic of Korea}. pp. \bibinfo{pages}{104--110}.
\newblock \URLprefix \url{https://aclanthology.org/2022.argmining-1.9}.
\bibitem[{Saadat-Yazdi et~al.(2023)Saadat-Yazdi, Pan and Kokciyan}]{saadat-yazdi-etal-2023-uncovering}
\bibinfo{author}{Saadat-Yazdi, A.}, \bibinfo{author}{Pan, J.Z.}, \bibinfo{author}{Kokciyan, N.}, \bibinfo{year}{2023}.
\newblock \bibinfo{title}{Uncovering implicit inferences for improved relational argument mining}, in: \bibinfo{booktitle}{Proceedings of the 17th Conference of the European Chapter of the Association for Computational Linguistics}, \bibinfo{publisher}{Association for Computational Linguistics}, \bibinfo{address}{Dubrovnik, Croatia}. pp. \bibinfo{pages}{2484--2495}.
\newblock \URLprefix \url{https://aclanthology.org/2023.eacl-main.182}.
\bibitem[{Saedi et~al.(2018)Saedi, Branco, Ant{\'o}nio~Rodrigues and Silva}]{saedi-etal-2018-wordnet}
\bibinfo{author}{Saedi, C.}, \bibinfo{author}{Branco, A.}, \bibinfo{author}{Ant{\'o}nio~Rodrigues, J.}, \bibinfo{author}{Silva, J.}, \bibinfo{year}{2018}.
\newblock \bibinfo{title}{{W}ord{N}et embeddings}, in: \bibinfo{booktitle}{Proceedings of the Third Workshop on Representation Learning for {NLP}}, \bibinfo{publisher}{Association for Computational Linguistics}, \bibinfo{address}{Melbourne, Australia}. pp. \bibinfo{pages}{122--131}.
\newblock \URLprefix \url{https://aclanthology.org/W18-3016}, \DOIprefix\doi{10.18653/v1/W18-3016}.
\bibitem[{Safavi and Koutra(2021)}]{safavi2021relational}
\bibinfo{author}{Safavi, T.}, \bibinfo{author}{Koutra, D.}, \bibinfo{year}{2021}.
\newblock \bibinfo{title}{Relational world knowledge representation in contextual language models: A review}, in: \bibinfo{booktitle}{Proceedings of the 2021 Conference on Empirical Methods in Natural Language Processing}, pp. \bibinfo{pages}{1053--1067}.
\bibitem[{Sap et~al.(2019)Sap, Le~Bras, Allaway, Bhagavatula, Lourie, Rashkin, Roof, Smith and Choi}]{sap2019atomic}
\bibinfo{author}{Sap, M.}, \bibinfo{author}{Le~Bras, R.}, \bibinfo{author}{Allaway, E.}, \bibinfo{author}{Bhagavatula, C.}, \bibinfo{author}{Lourie, N.}, \bibinfo{author}{Rashkin, H.}, \bibinfo{author}{Roof, B.}, \bibinfo{author}{Smith, N.A.}, \bibinfo{author}{Choi, Y.}, \bibinfo{year}{2019}.
\newblock \bibinfo{title}{Atomic: An atlas of machine commonsense for if-then reasoning}, in: \bibinfo{booktitle}{Proceedings of the AAAI conference on artificial intelligence}, pp. \bibinfo{pages}{3027--3035}.
\bibitem[{Sarker et~al.(2021)Sarker, Zhou, Eberhart and Hitzler}]{sarker2021neuro}
\bibinfo{author}{Sarker, M.K.}, \bibinfo{author}{Zhou, L.}, \bibinfo{author}{Eberhart, A.}, \bibinfo{author}{Hitzler, P.}, \bibinfo{year}{2021}.
\newblock \bibinfo{title}{Neuro-symbolic artificial intelligence}.
\newblock \bibinfo{journal}{AI Communications} \bibinfo{volume}{34}, \bibinfo{pages}{197--209}.
\bibitem[{Schlichtkrull et~al.(2018)Schlichtkrull, Kipf, Bloem, Van Den~Berg, Titov and Welling}]{schlichtkrull2018modeling}
\bibinfo{author}{Schlichtkrull, M.}, \bibinfo{author}{Kipf, T.N.}, \bibinfo{author}{Bloem, P.}, \bibinfo{author}{Van Den~Berg, R.}, \bibinfo{author}{Titov, I.}, \bibinfo{author}{Welling, M.}, \bibinfo{year}{2018}.
\newblock \bibinfo{title}{Modeling relational data with graph convolutional networks}, in: \bibinfo{booktitle}{The Semantic Web: 15th International Conference, ESWC 2018, Heraklion, Crete, Greece, June 3--7, 2018, Proceedings 15}, \bibinfo{organization}{Springer}. pp. \bibinfo{pages}{593--607}.
\bibitem[{Schneider et~al.(2022)Schneider, Schopf, Vladika, Galkin, Simperl and Matthes}]{schneider-etal-2022-decade}
\bibinfo{author}{Schneider, P.}, \bibinfo{author}{Schopf, T.}, \bibinfo{author}{Vladika, J.}, \bibinfo{author}{Galkin, M.}, \bibinfo{author}{Simperl, E.}, \bibinfo{author}{Matthes, F.}, \bibinfo{year}{2022}.
\newblock \bibinfo{title}{A decade of knowledge graphs in natural language processing: A survey}, in: \bibinfo{booktitle}{Proceedings of the 2nd Conference of the Asia-Pacific Chapter of the Association for Computational Linguistics and the 12th International Joint Conference on Natural Language Processing (Volume 1: Long Papers)}, \bibinfo{publisher}{Association for Computational Linguistics}, \bibinfo{address}{Online only}. pp. \bibinfo{pages}{601--614}.
\newblock \URLprefix \url{https://aclanthology.org/2022.aacl-main.46}.
\bibitem[{Sebastiani(2002)}]{sebastiani2002machine}
\bibinfo{author}{Sebastiani, F.}, \bibinfo{year}{2002}.
\newblock \bibinfo{title}{Machine learning in automated text categorization}.
\newblock \bibinfo{journal}{ACM computing surveys (CSUR)} \bibinfo{volume}{34}, \bibinfo{pages}{1--47}.
\bibitem[{Sennrich and Haddow(2016)}]{sennrich-haddow-2016-linguistic}
\bibinfo{author}{Sennrich, R.}, \bibinfo{author}{Haddow, B.}, \bibinfo{year}{2016}.
\newblock \bibinfo{title}{Linguistic input features improve neural machine translation}, in: \bibinfo{booktitle}{Proceedings of the First Conference on Machine Translation: Volume 1, Research Papers}, \bibinfo{publisher}{Association for Computational Linguistics}, \bibinfo{address}{Berlin, Germany}. pp. \bibinfo{pages}{83--91}.
\newblock \URLprefix \url{https://aclanthology.org/W16-2209}, \DOIprefix\doi{10.18653/v1/W16-2209}.
\bibitem[{Shi and Weninger(2016)}]{shi2016fact}
\bibinfo{author}{Shi, B.}, \bibinfo{author}{Weninger, T.}, \bibinfo{year}{2016}.
\newblock \bibinfo{title}{Fact checking in heterogeneous information networks}, in: \bibinfo{booktitle}{Proceedings of the 25th International Conference Companion on World Wide Web}, pp. \bibinfo{pages}{101--102}.
\bibitem[{Shiralkar et~al.(2017)Shiralkar, Flammini, Menczer and Ciampaglia}]{shiralkar2017finding}
\bibinfo{author}{Shiralkar, P.}, \bibinfo{author}{Flammini, A.}, \bibinfo{author}{Menczer, F.}, \bibinfo{author}{Ciampaglia, G.L.}, \bibinfo{year}{2017}.
\newblock \bibinfo{title}{Finding streams in knowledge graphs to support fact checking}, in: \bibinfo{booktitle}{2017 IEEE International Conference on Data Mining (ICDM)}, \bibinfo{organization}{IEEE}. pp. \bibinfo{pages}{859--864}.
\bibitem[{Si et~al.(2021)Si, Zhou, Li, Shi and He}]{si-etal-2021-topic}
\bibinfo{author}{Si, J.}, \bibinfo{author}{Zhou, D.}, \bibinfo{author}{Li, T.}, \bibinfo{author}{Shi, X.}, \bibinfo{author}{He, Y.}, \bibinfo{year}{2021}.
\newblock \bibinfo{title}{Topic-aware evidence reasoning and stance-aware aggregation for fact verification}, in: \bibinfo{booktitle}{Proceedings of the 59th Annual Meeting of the Association for Computational Linguistics and the 11th International Joint Conference on Natural Language Processing (Volume 1: Long Papers)}, \bibinfo{publisher}{Association for Computational Linguistics}, \bibinfo{address}{Online}. pp. \bibinfo{pages}{1612--1622}.
\newblock \URLprefix \url{https://aclanthology.org/2021.acl-long.128}, \DOIprefix\doi{10.18653/v1/2021.acl-long.128}.
\bibitem[{{\v{S}}krlj et~al.(2021){\v{S}}krlj, Martinc, Kralj, Lavra{\v{c}} and Pollak}]{vskrlj2021tax2vec}
\bibinfo{author}{{\v{S}}krlj, B.}, \bibinfo{author}{Martinc, M.}, \bibinfo{author}{Kralj, J.}, \bibinfo{author}{Lavra{\v{c}}, N.}, \bibinfo{author}{Pollak, S.}, \bibinfo{year}{2021}.
\newblock \bibinfo{title}{tax2vec: Constructing interpretable features from taxonomies for short text classification}.
\newblock \bibinfo{journal}{Computer Speech \& Language} \bibinfo{volume}{65}, \bibinfo{pages}{101104}.
\bibitem[{Speer et~al.(2017)Speer, Chin and Havasi}]{speer2017conceptnet}
\bibinfo{author}{Speer, R.}, \bibinfo{author}{Chin, J.}, \bibinfo{author}{Havasi, C.}, \bibinfo{year}{2017}.
\newblock \bibinfo{title}{Conceptnet 5.5: An open multilingual graph of general knowledge}, in: \bibinfo{booktitle}{Proceedings of the AAAI conference on artificial intelligence}.
\bibitem[{Stasaski and Hearst(2017)}]{stasaski-hearst-2017-multiple}
\bibinfo{author}{Stasaski, K.}, \bibinfo{author}{Hearst, M.A.}, \bibinfo{year}{2017}.
\newblock \bibinfo{title}{Multiple choice question generation utilizing an ontology}, in: \bibinfo{booktitle}{Proceedings of the 12th Workshop on Innovative Use of {NLP} for Building Educational Applications}, \bibinfo{publisher}{Association for Computational Linguistics}, \bibinfo{address}{Copenhagen, Denmark}. pp. \bibinfo{pages}{303--312}.
\newblock \URLprefix \url{https://aclanthology.org/W17-5034}, \DOIprefix\doi{10.18653/v1/W17-5034}.
\bibitem[{Suchanek et~al.(2007)Suchanek, Kasneci and Weikum}]{suchanek2007yago}
\bibinfo{author}{Suchanek, F.M.}, \bibinfo{author}{Kasneci, G.}, \bibinfo{author}{Weikum, G.}, \bibinfo{year}{2007}.
\newblock \bibinfo{title}{Yago: a core of semantic knowledge}, in: \bibinfo{booktitle}{Proceedings of the 16th international conference on World Wide Web}, pp. \bibinfo{pages}{697--706}.
\bibitem[{Sun et~al.(2019)Sun, Deng, Nie and Tang}]{sun2019rotate}
\bibinfo{author}{Sun, Z.}, \bibinfo{author}{Deng, Z.H.}, \bibinfo{author}{Nie, J.Y.}, \bibinfo{author}{Tang, J.}, \bibinfo{year}{2019}.
\newblock \bibinfo{title}{Rotate: Knowledge graph embedding by relational rotation in complex space}.
\newblock \bibinfo{journal}{arXiv preprint arXiv:1902.10197} .
\bibitem[{Tedeschi et~al.(2021)Tedeschi, Maiorca, Campolungo, Cecconi and Navigli}]{tedeschi2021wikineural}
\bibinfo{author}{Tedeschi, S.}, \bibinfo{author}{Maiorca, V.}, \bibinfo{author}{Campolungo, N.}, \bibinfo{author}{Cecconi, F.}, \bibinfo{author}{Navigli, R.}, \bibinfo{year}{2021}.
\newblock \bibinfo{title}{Wikineural: Combined neural and knowledge-based silver data creation for multilingual ner}, in: \bibinfo{booktitle}{Findings of the Association for Computational Linguistics: EMNLP 2021}, pp. \bibinfo{pages}{2521--2533}.
\bibitem[{Vaswani et~al.(2017)Vaswani, Shazeer, Parmar, Uszkoreit, Jones, Gomez, Kaiser and Polosukhin}]{vaswani2017attention}
\bibinfo{author}{Vaswani, A.}, \bibinfo{author}{Shazeer, N.}, \bibinfo{author}{Parmar, N.}, \bibinfo{author}{Uszkoreit, J.}, \bibinfo{author}{Jones, L.}, \bibinfo{author}{Gomez, A.N.}, \bibinfo{author}{Kaiser, {\L}.}, \bibinfo{author}{Polosukhin, I.}, \bibinfo{year}{2017}.
\newblock \bibinfo{title}{Attention is all you need}.
\newblock \bibinfo{journal}{Advances in neural information processing systems} \bibinfo{volume}{30}.
\bibitem[{Velickovic et~al.(2017)Velickovic, Cucurull, Casanova, Romero, Lio, Bengio et~al.}]{velickovic2017graph}
\bibinfo{author}{Velickovic, P.}, \bibinfo{author}{Cucurull, G.}, \bibinfo{author}{Casanova, A.}, \bibinfo{author}{Romero, A.}, \bibinfo{author}{Lio, P.}, \bibinfo{author}{Bengio, Y.}, et~al., \bibinfo{year}{2017}.
\newblock \bibinfo{title}{Graph attention networks}.
\newblock \bibinfo{journal}{stat} \bibinfo{volume}{1050}, \bibinfo{pages}{10--48550}.
\bibitem[{Vrande{\v{c}}i{\'c} and Kr{\"o}tzsch(2014)}]{vrandevcic2014wikidata}
\bibinfo{author}{Vrande{\v{c}}i{\'c}, D.}, \bibinfo{author}{Kr{\"o}tzsch, M.}, \bibinfo{year}{2014}.
\newblock \bibinfo{title}{Wikidata: a free collaborative knowledgebase}.
\newblock \bibinfo{journal}{Communications of the ACM} \bibinfo{volume}{57}, \bibinfo{pages}{78--85}.
\bibitem[{Wang et~al.(2020a)Wang, Liu, Zheng, Qiu and Huang}]{wang-etal-2020-heterogeneous}
\bibinfo{author}{Wang, D.}, \bibinfo{author}{Liu, P.}, \bibinfo{author}{Zheng, Y.}, \bibinfo{author}{Qiu, X.}, \bibinfo{author}{Huang, X.}, \bibinfo{year}{2020}a.
\newblock \bibinfo{title}{Heterogeneous graph neural networks for extractive document summarization}, in: \bibinfo{booktitle}{Proceedings of the 58th Annual Meeting of the Association for Computational Linguistics}, \bibinfo{publisher}{Association for Computational Linguistics}, \bibinfo{address}{Online}. pp. \bibinfo{pages}{6209--6219}.
\newblock \URLprefix \url{https://aclanthology.org/2020.acl-main.553}, \DOIprefix\doi{10.18653/v1/2020.acl-main.553}.
\bibitem[{Wang et~al.(2022a)Wang, Li, Lai and Jiang}]{Wang2022KATSumKA}
\bibinfo{author}{Wang, G.H.}, \bibinfo{author}{Li, W.}, \bibinfo{author}{Lai, E.M.K.}, \bibinfo{author}{Jiang, J.}, \bibinfo{year}{2022}a.
\newblock \bibinfo{title}{Katsum: Knowledge-aware abstractive text summarization}.
\newblock \bibinfo{journal}{ArXiv} \bibinfo{volume}{abs/2212.03371}.
\newblock \URLprefix \url{https://api.semanticscholar.org/CorpusID:254366408}.
\bibitem[{Wang et~al.(2020b)Wang, Wang, Qi and Zheng}]{wang2020richpedia}
\bibinfo{author}{Wang, M.}, \bibinfo{author}{Wang, H.}, \bibinfo{author}{Qi, G.}, \bibinfo{author}{Zheng, Q.}, \bibinfo{year}{2020}b.
\newblock \bibinfo{title}{Richpedia: a large-scale, comprehensive multi-modal knowledge graph}.
\newblock \bibinfo{journal}{Big Data Research} \bibinfo{volume}{22}, \bibinfo{pages}{100159}.
\bibitem[{Wang et~al.(2021)Wang, Gao, Zhu, Zhang, Liu, Li and Tang}]{wang2021kepler}
\bibinfo{author}{Wang, X.}, \bibinfo{author}{Gao, T.}, \bibinfo{author}{Zhu, Z.}, \bibinfo{author}{Zhang, Z.}, \bibinfo{author}{Liu, Z.}, \bibinfo{author}{Li, J.}, \bibinfo{author}{Tang, J.}, \bibinfo{year}{2021}.
\newblock \bibinfo{title}{Kepler: A unified model for knowledge embedding and pre-trained language representation}.
\newblock \bibinfo{journal}{Transactions of the Association for Computational Linguistics} \bibinfo{volume}{9}, \bibinfo{pages}{176--194}.
\bibitem[{Wang et~al.(2022b)Wang, Shen, Cai, Wang, Wang, Xie, Huang, Lu, Zhuang, Tu, Lu and Jiang}]{wang-etal-2022-damo}
\bibinfo{author}{Wang, X.}, \bibinfo{author}{Shen, Y.}, \bibinfo{author}{Cai, J.}, \bibinfo{author}{Wang, T.}, \bibinfo{author}{Wang, X.}, \bibinfo{author}{Xie, P.}, \bibinfo{author}{Huang, F.}, \bibinfo{author}{Lu, W.}, \bibinfo{author}{Zhuang, Y.}, \bibinfo{author}{Tu, K.}, \bibinfo{author}{Lu, W.}, \bibinfo{author}{Jiang, Y.}, \bibinfo{year}{2022}b.
\newblock \bibinfo{title}{{DAMO}-{NLP} at {S}em{E}val-2022 task 11: A knowledge-based system for multilingual named entity recognition}, in: \bibinfo{booktitle}{Proceedings of the 16th International Workshop on Semantic Evaluation (SemEval-2022)}, \bibinfo{publisher}{Association for Computational Linguistics}, \bibinfo{address}{Seattle, United States}. pp. \bibinfo{pages}{1457--1468}.
\newblock \URLprefix \url{https://aclanthology.org/2022.semeval-1.200}, \DOIprefix\doi{10.18653/v1/2022.semeval-1.200}.
\bibitem[{Wang et~al.(2020c)Wang, Sun, Ma, Gao and Xu}]{wang2020ernie}
\bibinfo{author}{Wang, Y.}, \bibinfo{author}{Sun, Y.}, \bibinfo{author}{Ma, Z.}, \bibinfo{author}{Gao, L.}, \bibinfo{author}{Xu, Y.}, \bibinfo{year}{2020}c.
\newblock \bibinfo{title}{An ernie-based joint model for chinese named entity recognition}.
\newblock \bibinfo{journal}{Applied Sciences} \bibinfo{volume}{10}, \bibinfo{pages}{5711}.
\bibitem[{Wang et~al.(2010)Wang, Huang, Li, Liu, Shao, Wang, Wang, Wang, Wu, Xiao et~al.}]{wang2010probase}
\bibinfo{author}{Wang, Z.}, \bibinfo{author}{Huang, J.}, \bibinfo{author}{Li, H.}, \bibinfo{author}{Liu, B.}, \bibinfo{author}{Shao, B.}, \bibinfo{author}{Wang, H.}, \bibinfo{author}{Wang, J.}, \bibinfo{author}{Wang, Y.}, \bibinfo{author}{Wu, W.}, \bibinfo{author}{Xiao, J.}, et~al., \bibinfo{year}{2010}.
\newblock \bibinfo{title}{Probase: a universal knowledge base for semantic search}.
\newblock \bibinfo{journal}{Microsoft Research Asia} .
\bibitem[{Wu et~al.(2012)Wu, Li, Wang and Zhu}]{wu2012probase}
\bibinfo{author}{Wu, W.}, \bibinfo{author}{Li, H.}, \bibinfo{author}{Wang, H.}, \bibinfo{author}{Zhu, K.Q.}, \bibinfo{year}{2012}.
\newblock \bibinfo{title}{Probase: A probabilistic taxonomy for text understanding}, in: \bibinfo{booktitle}{Proceedings of the 2012 ACM SIGMOD international conference on management of data}, pp. \bibinfo{pages}{481--492}.
\bibitem[{Xiong et~al.(2019)Xiong, Du, Wang and Stoyanov}]{xiong2019pretrained}
\bibinfo{author}{Xiong, W.}, \bibinfo{author}{Du, J.}, \bibinfo{author}{Wang, W.Y.}, \bibinfo{author}{Stoyanov, V.}, \bibinfo{year}{2019}.
\newblock \bibinfo{title}{Pretrained encyclopedia: Weakly supervised knowledge-pretrained language model}.
\newblock \bibinfo{journal}{arXiv preprint arXiv:1912.09637} .
\bibitem[{Xu et~al.(2020)Xu, Gan, Cheng and Liu}]{xu-etal-2020-discourse}
\bibinfo{author}{Xu, J.}, \bibinfo{author}{Gan, Z.}, \bibinfo{author}{Cheng, Y.}, \bibinfo{author}{Liu, J.}, \bibinfo{year}{2020}.
\newblock \bibinfo{title}{Discourse-aware neural extractive text summarization}, in: \bibinfo{booktitle}{Proceedings of the 58th Annual Meeting of the Association for Computational Linguistics}, \bibinfo{publisher}{Association for Computational Linguistics}, \bibinfo{address}{Online}. pp. \bibinfo{pages}{5021--5031}.
\newblock \URLprefix \url{https://aclanthology.org/2020.acl-main.451}, \DOIprefix\doi{10.18653/v1/2020.acl-main.451}.
\bibitem[{Xu et~al.(2022)Xu, Wu, Liu, Wu and Wang}]{xu2022evidence}
\bibinfo{author}{Xu, W.}, \bibinfo{author}{Wu, J.}, \bibinfo{author}{Liu, Q.}, \bibinfo{author}{Wu, S.}, \bibinfo{author}{Wang, L.}, \bibinfo{year}{2022}.
\newblock \bibinfo{title}{Evidence-aware fake news detection with graph neural networks}, in: \bibinfo{booktitle}{Proceedings of the ACM Web Conference 2022}, pp. \bibinfo{pages}{2501--2510}.
\bibitem[{Yang et~al.(2019)Yang, Dai, Yang, Carbonell, Salakhutdinov and Le}]{yang2019xlnet}
\bibinfo{author}{Yang, Z.}, \bibinfo{author}{Dai, Z.}, \bibinfo{author}{Yang, Y.}, \bibinfo{author}{Carbonell, J.}, \bibinfo{author}{Salakhutdinov, R.R.}, \bibinfo{author}{Le, Q.V.}, \bibinfo{year}{2019}.
\newblock \bibinfo{title}{Xlnet: Generalized autoregressive pretraining for language understanding}.
\newblock \bibinfo{journal}{Advances in neural information processing systems} \bibinfo{volume}{32}.
\bibitem[{Yang et~al.(2023)Yang, Xu, Hu and Dong}]{YANG2023103245}
\bibinfo{author}{Yang, Z.}, \bibinfo{author}{Xu, Y.}, \bibinfo{author}{Hu, J.}, \bibinfo{author}{Dong, S.}, \bibinfo{year}{2023}.
\newblock \bibinfo{title}{Generating knowledge aware explanation for natural language inference}.
\newblock \bibinfo{journal}{Information Processing \& Management} \bibinfo{volume}{60}, \bibinfo{pages}{103245}.
\newblock \URLprefix \url{https://www.sciencedirect.com/science/article/pii/S0306457322003466}, \DOIprefix\doi{https://doi.org/10.1016/j.ipm.2022.103245}.
\bibitem[{Yao et~al.(2019)Yao, Mao and Luo}]{yao2019graph}
\bibinfo{author}{Yao, L.}, \bibinfo{author}{Mao, C.}, \bibinfo{author}{Luo, Y.}, \bibinfo{year}{2019}.
\newblock \bibinfo{title}{Graph convolutional networks for text classification}, in: \bibinfo{booktitle}{Proceedings of the AAAI conference on artificial intelligence}, pp. \bibinfo{pages}{7370--7377}.
\bibitem[{Yasunaga et~al.(2021)Yasunaga, Ren, Bosselut, Liang and Leskovec}]{yasunaga-etal-2021-qa}
\bibinfo{author}{Yasunaga, M.}, \bibinfo{author}{Ren, H.}, \bibinfo{author}{Bosselut, A.}, \bibinfo{author}{Liang, P.}, \bibinfo{author}{Leskovec, J.}, \bibinfo{year}{2021}.
\newblock \bibinfo{title}{{QA}-{GNN}: Reasoning with language models and knowledge graphs for question answering}, in: \bibinfo{booktitle}{Proceedings of the 2021 Conference of the North American Chapter of the Association for Computational Linguistics: Human Language Technologies}, \bibinfo{publisher}{Association for Computational Linguistics}, \bibinfo{address}{Online}. pp. \bibinfo{pages}{535--546}.
\newblock \URLprefix \url{https://aclanthology.org/2021.naacl-main.45}, \DOIprefix\doi{10.18653/v1/2021.naacl-main.45}.
\bibitem[{Yin et~al.(2022)Yin, Dong, Cheng, Liu, Chang, Wei and Gao}]{yin2022survey}
\bibinfo{author}{Yin, D.}, \bibinfo{author}{Dong, L.}, \bibinfo{author}{Cheng, H.}, \bibinfo{author}{Liu, X.}, \bibinfo{author}{Chang, K.W.}, \bibinfo{author}{Wei, F.}, \bibinfo{author}{Gao, J.}, \bibinfo{year}{2022}.
\newblock \bibinfo{title}{A survey of knowledge-intensive nlp with pre-trained language models}.
\newblock \bibinfo{journal}{arXiv preprint arXiv:2202.08772} .
\bibitem[{Yu et~al.(2023a)Yu, Yang, Liu, Wang and Pan}]{YU2023105}
\bibinfo{author}{Yu, D.}, \bibinfo{author}{Yang, B.}, \bibinfo{author}{Liu, D.}, \bibinfo{author}{Wang, H.}, \bibinfo{author}{Pan, S.}, \bibinfo{year}{2023}a.
\newblock \bibinfo{title}{A survey on neural-symbolic learning systems}.
\newblock \bibinfo{journal}{Neural Networks} \bibinfo{volume}{166}, \bibinfo{pages}{105--126}.
\newblock \URLprefix \url{https://www.sciencedirect.com/science/article/pii/S0893608023003398}, \DOIprefix\doi{https://doi.org/10.1016/j.neunet.2023.06.028}.
\bibitem[{Yu et~al.(2022a)Yu, Zhu, Yang and Zeng}]{yu2022jaket}
\bibinfo{author}{Yu, D.}, \bibinfo{author}{Zhu, C.}, \bibinfo{author}{Yang, Y.}, \bibinfo{author}{Zeng, M.}, \bibinfo{year}{2022}a.
\newblock \bibinfo{title}{Jaket: Joint pre-training of knowledge graph and language understanding}, in: \bibinfo{booktitle}{Proceedings of the AAAI Conference on Artificial Intelligence}, pp. \bibinfo{pages}{11630--11638}.
\bibitem[{Yu et~al.(2023b)Yu, Zhang and Wang}]{yu2023nature}
\bibinfo{author}{Yu, F.}, \bibinfo{author}{Zhang, H.}, \bibinfo{author}{Wang, B.}, \bibinfo{year}{2023}b.
\newblock \bibinfo{title}{Nature language reasoning, a survey}.
\newblock \bibinfo{journal}{arXiv preprint arXiv:2303.14725} .
\bibitem[{Yu et~al.(2022b)Yu, Zhu, Li, Hu, Wang, Ji and Jiang}]{yu2022survey}
\bibinfo{author}{Yu, W.}, \bibinfo{author}{Zhu, C.}, \bibinfo{author}{Li, Z.}, \bibinfo{author}{Hu, Z.}, \bibinfo{author}{Wang, Q.}, \bibinfo{author}{Ji, H.}, \bibinfo{author}{Jiang, M.}, \bibinfo{year}{2022}b.
\newblock \bibinfo{title}{A survey of knowledge-enhanced text generation}.
\newblock \bibinfo{journal}{ACM Computing Surveys} \bibinfo{volume}{54}, \bibinfo{pages}{1--38}.
\bibitem[{Zaib et~al.(2022)Zaib, Zhang, Sheng, Mahmood and Zhang}]{zaib2022conversational}
\bibinfo{author}{Zaib, M.}, \bibinfo{author}{Zhang, W.E.}, \bibinfo{author}{Sheng, Q.Z.}, \bibinfo{author}{Mahmood, A.}, \bibinfo{author}{Zhang, Y.}, \bibinfo{year}{2022}.
\newblock \bibinfo{title}{Conversational question answering: A survey}.
\newblock \bibinfo{journal}{Knowledge and Information Systems} \bibinfo{volume}{64}, \bibinfo{pages}{3151--3195}.
\bibitem[{Zeng et~al.(2021)Zeng, Abumansour and Zubiaga}]{zeng2021automated}
\bibinfo{author}{Zeng, X.}, \bibinfo{author}{Abumansour, A.S.}, \bibinfo{author}{Zubiaga, A.}, \bibinfo{year}{2021}.
\newblock \bibinfo{title}{Automated fact-checking: A survey}.
\newblock \bibinfo{journal}{Language and Linguistics Compass} \bibinfo{volume}{15}, \bibinfo{pages}{e12438}.
\bibitem[{Zhang et~al.(2020a)Zhang, Khashabi, Song and Roth}]{zhang2020transomcs}
\bibinfo{author}{Zhang, H.}, \bibinfo{author}{Khashabi, D.}, \bibinfo{author}{Song, Y.}, \bibinfo{author}{Roth, D.}, \bibinfo{year}{2020}a.
\newblock \bibinfo{title}{Transomcs: From linguistic graphs to commonsense knowledge}.
\newblock \bibinfo{journal}{arXiv preprint arXiv:2005.00206} .
\bibitem[{Zhang et~al.(2020b)Zhang, Liu, Pan, Song and Leung}]{zhang2020aser}
\bibinfo{author}{Zhang, H.}, \bibinfo{author}{Liu, X.}, \bibinfo{author}{Pan, H.}, \bibinfo{author}{Song, Y.}, \bibinfo{author}{Leung, C.W.K.}, \bibinfo{year}{2020}b.
\newblock \bibinfo{title}{Aser: A large-scale eventuality knowledge graph}, in: \bibinfo{booktitle}{Proceedings of the web conference 2020}, pp. \bibinfo{pages}{201--211}.
\bibitem[{Zhang et~al.(2020c)Zhang, Liu, Xiong and Liu}]{zhang-etal-2020-grounded}
\bibinfo{author}{Zhang, H.}, \bibinfo{author}{Liu, Z.}, \bibinfo{author}{Xiong, C.}, \bibinfo{author}{Liu, Z.}, \bibinfo{year}{2020}c.
\newblock \bibinfo{title}{Grounded conversation generation as guided traverses in commonsense knowledge graphs}, in: \bibinfo{booktitle}{Proceedings of the 58th Annual Meeting of the Association for Computational Linguistics}, \bibinfo{publisher}{Association for Computational Linguistics}, \bibinfo{address}{Online}. pp. \bibinfo{pages}{2031--2043}.
\newblock \URLprefix \url{https://aclanthology.org/2020.acl-main.184}, \DOIprefix\doi{10.18653/v1/2020.acl-main.184}.
\bibitem[{Zhang et~al.(2023a)Zhang, Chen, Fang and Chen}]{ZHANG2023103297}
\bibinfo{author}{Zhang, Q.}, \bibinfo{author}{Chen, S.}, \bibinfo{author}{Fang, M.}, \bibinfo{author}{Chen, X.}, \bibinfo{year}{2023}a.
\newblock \bibinfo{title}{Joint reasoning with knowledge subgraphs for multiple choice question answering}.
\newblock \bibinfo{journal}{Information Processing \& Management} \bibinfo{volume}{60}, \bibinfo{pages}{103297}.
\newblock \URLprefix \url{https://www.sciencedirect.com/science/article/pii/S0306457323000341}, \DOIprefix\doi{https://doi.org/10.1016/j.ipm.2023.103297}.
\bibitem[{Zhang et~al.(2019a)Zhang, Tay, Yao and Liu}]{zhang2019quaternion}
\bibinfo{author}{Zhang, S.}, \bibinfo{author}{Tay, Y.}, \bibinfo{author}{Yao, L.}, \bibinfo{author}{Liu, Q.}, \bibinfo{year}{2019}a.
\newblock \bibinfo{title}{Quaternion knowledge graph embeddings}.
\newblock \bibinfo{journal}{Advances in neural information processing systems} \bibinfo{volume}{32}.
\bibitem[{Zhang et~al.(2022)Zhang, Bosselut, Yasunaga, Ren, Liang, Manning and Leskovec}]{zhang2022greaselm}
\bibinfo{author}{Zhang, X.}, \bibinfo{author}{Bosselut, A.}, \bibinfo{author}{Yasunaga, M.}, \bibinfo{author}{Ren, H.}, \bibinfo{author}{Liang, P.}, \bibinfo{author}{Manning, C.D.}, \bibinfo{author}{Leskovec, J.}, \bibinfo{year}{2022}.
\newblock \bibinfo{title}{Greaselm: Graph reasoning enhanced language models for question answering}.
\newblock \bibinfo{journal}{arXiv preprint arXiv:2201.08860} .
\bibitem[{Zhang et~al.(2023b)Zhang, Li, Cui, Cai, Liu, Fu, Huang, Zhao, Zhang, Chen et~al.}]{zhang2023siren}
\bibinfo{author}{Zhang, Y.}, \bibinfo{author}{Li, Y.}, \bibinfo{author}{Cui, L.}, \bibinfo{author}{Cai, D.}, \bibinfo{author}{Liu, L.}, \bibinfo{author}{Fu, T.}, \bibinfo{author}{Huang, X.}, \bibinfo{author}{Zhao, E.}, \bibinfo{author}{Zhang, Y.}, \bibinfo{author}{Chen, Y.}, et~al., \bibinfo{year}{2023}b.
\newblock \bibinfo{title}{Siren's song in the ai ocean: A survey on hallucination in large language models}.
\newblock \bibinfo{journal}{arXiv preprint arXiv:2309.01219} .
\bibitem[{Zhang et~al.(2019b)Zhang, Han, Liu, Jiang, Sun and Liu}]{zhang-etal-2019-ernie}
\bibinfo{author}{Zhang, Z.}, \bibinfo{author}{Han, X.}, \bibinfo{author}{Liu, Z.}, \bibinfo{author}{Jiang, X.}, \bibinfo{author}{Sun, M.}, \bibinfo{author}{Liu, Q.}, \bibinfo{year}{2019}b.
\newblock \bibinfo{title}{{ERNIE}: Enhanced language representation with informative entities}, in: \bibinfo{booktitle}{Proceedings of the 57th Annual Meeting of the Association for Computational Linguistics}, \bibinfo{publisher}{Association for Computational Linguistics}, \bibinfo{address}{Florence, Italy}. pp. \bibinfo{pages}{1441--1451}.
\newblock \URLprefix \url{https://aclanthology.org/P19-1139}, \DOIprefix\doi{10.18653/v1/P19-1139}.
\bibitem[{Zhong et~al.(2020)Zhong, Xu, Tang, Xu, Duan, Zhou, Wang and Yin}]{zhong-etal-2020-reasoning}
\bibinfo{author}{Zhong, W.}, \bibinfo{author}{Xu, J.}, \bibinfo{author}{Tang, D.}, \bibinfo{author}{Xu, Z.}, \bibinfo{author}{Duan, N.}, \bibinfo{author}{Zhou, M.}, \bibinfo{author}{Wang, J.}, \bibinfo{author}{Yin, J.}, \bibinfo{year}{2020}.
\newblock \bibinfo{title}{Reasoning over semantic-level graph for fact checking}, in: \bibinfo{booktitle}{Proceedings of the 58th Annual Meeting of the Association for Computational Linguistics}, \bibinfo{publisher}{Association for Computational Linguistics}, \bibinfo{address}{Online}. pp. \bibinfo{pages}{6170--6180}.
\newblock \URLprefix \url{https://aclanthology.org/2020.acl-main.549}, \DOIprefix\doi{10.18653/v1/2020.acl-main.549}.
\bibitem[{Zhou et~al.(2018)Zhou, Young, Huang, Zhao, Xu and Zhu}]{zhou2018commonsense}
\bibinfo{author}{Zhou, H.}, \bibinfo{author}{Young, T.}, \bibinfo{author}{Huang, M.}, \bibinfo{author}{Zhao, H.}, \bibinfo{author}{Xu, J.}, \bibinfo{author}{Zhu, X.}, \bibinfo{year}{2018}.
\newblock \bibinfo{title}{Commonsense knowledge aware conversation generation with graph attention.}, in: \bibinfo{booktitle}{IJCAI}, pp. \bibinfo{pages}{4623--4629}.
\bibitem[{Zhou et~al.(2019)Zhou, Han, Yang, Liu, Wang, Li and Sun}]{zhou-etal-2019-gear}
\bibinfo{author}{Zhou, J.}, \bibinfo{author}{Han, X.}, \bibinfo{author}{Yang, C.}, \bibinfo{author}{Liu, Z.}, \bibinfo{author}{Wang, L.}, \bibinfo{author}{Li, C.}, \bibinfo{author}{Sun, M.}, \bibinfo{year}{2019}.
\newblock \bibinfo{title}{{GEAR}: Graph-based evidence aggregating and reasoning for fact verification}, in: \bibinfo{booktitle}{Proceedings of the 57th Annual Meeting of the Association for Computational Linguistics}, \bibinfo{publisher}{Association for Computational Linguistics}, \bibinfo{address}{Florence, Italy}. pp. \bibinfo{pages}{892--901}.
\newblock \URLprefix \url{https://aclanthology.org/P19-1085}, \DOIprefix\doi{10.18653/v1/P19-1085}.
\bibitem[{Zhou et~al.(2020)Zhou, Lee, Selvam, Lee, Lin and Ren}]{zhou2020pre}
\bibinfo{author}{Zhou, W.}, \bibinfo{author}{Lee, D.H.}, \bibinfo{author}{Selvam, R.K.}, \bibinfo{author}{Lee, S.}, \bibinfo{author}{Lin, B.Y.}, \bibinfo{author}{Ren, X.}, \bibinfo{year}{2020}.
\newblock \bibinfo{title}{Pre-training text-to-text transformers for concept-centric common sense}.
\newblock \bibinfo{journal}{arXiv preprint arXiv:2011.07956} .
\bibitem[{Zhu et~al.(2023)Zhu, Xu, Ren, Lin, Jiang and Yu}]{zhu2023knowledge}
\bibinfo{author}{Zhu, C.}, \bibinfo{author}{Xu, Y.}, \bibinfo{author}{Ren, X.}, \bibinfo{author}{Lin, B.Y.}, \bibinfo{author}{Jiang, M.}, \bibinfo{author}{Yu, W.}, \bibinfo{year}{2023}.
\newblock \bibinfo{title}{Knowledge-augmented methods for natural language processing}, in: \bibinfo{booktitle}{Proceedings of the Sixteenth ACM International Conference on Web Search and Data Mining}, pp. \bibinfo{pages}{1228--1231}.
\bibitem[{Zhu et~al.(2017)Zhu, Zhou, Xu, Liu, Tan et~al.}]{zhu2017intelligent}
\bibinfo{author}{Zhu, Y.}, \bibinfo{author}{Zhou, W.}, \bibinfo{author}{Xu, Y.}, \bibinfo{author}{Liu, J.}, \bibinfo{author}{Tan, Y.}, et~al., \bibinfo{year}{2017}.
\newblock \bibinfo{title}{Intelligent learning for knowledge graph towards geological data}.
\newblock \bibinfo{journal}{Scientific Programming} \bibinfo{volume}{2017}.
\bibitem[{Zouhar et~al.(2022)Zouhar, Mosbach, Biswas and Klakow}]{zouhar2022artefact}
\bibinfo{author}{Zouhar, V.}, \bibinfo{author}{Mosbach, M.}, \bibinfo{author}{Biswas, D.}, \bibinfo{author}{Klakow, D.}, \bibinfo{year}{2022}.
\newblock \bibinfo{title}{Artefact retrieval: Overview of nlp models with knowledge base access}.
\newblock \bibinfo{journal}{arXiv preprint arXiv:2201.09651} .

\end{thebibliography}

\end{document}